%% file: main.tex
\documentclass[conference]{IEEEtran}
\usepackage{cite}
\usepackage{graphicx,epstopdf,url}
\usepackage{subfigure}
\usepackage{verbatim}
\usepackage{textcomp}
\usepackage{slashbox}
\usepackage{amsmath}
\usepackage{amssymb}
\usepackage{multirow}

\usepackage{xcolor}
\usepackage{soul}
\usepackage{tabularx}
\usepackage{makecell}
\usepackage{wrapfig}

\usepackage{booktabs}
\usepackage[utf8]{inputenc}  
\usepackage{indentfirst}
\usepackage[all]{xy}
\usepackage{bbm}
\usepackage{algorithm}
\usepackage{algpseudocode}
\usepackage{float}
\usepackage{graphicx}
\usepackage{diagbox}
\newcommand{\tabincell}[2]{\begin{tabular}{@{}#1@{}}#2\end{tabular}}
\def\BibTeX{{\rm B\kern-.05em{\sc i\kern-.025em b}\kern-.08em
    T\kern-.1667em\lower.7ex\hbox{E}\kern-.125emX}}
\begin{document}
%Here goes the title
\title{HgbNet: predicting hemoglobin level/anemia degree from EHR data}

\author{
    \IEEEauthorblockN{Zhuo ZHI$^{1}$,  Moe Elbadawi$^{2}$, Adam Daneshmend$^{3}$, Mine Orlu$^{2}$, Abdul Basit$^{2}$, \\ Andreas Demosthenous$^{1}$, \textit{Fellow, IEEE} and  Miguel Rodrigues$^{1}$, \textit{Fellow, IEEE}\\
   \IEEEauthorblockA{
    $^{1}$Department of Electronic and Electrical Engineering, University College London, London, UK. \\$^{2}$UCL School of Pharmacy, University College London, London, UK \\$^{3}$Imperial College Healthcare NHS Trust, London, UK}
    \\ \textcolor{red}{This work has been submitted to the IEEE for possible publication.}
    \\ \textcolor{red}{Copyright may be transferred without notice, after which this version may no longer be accessible}
    }
}
\maketitle

% \begin{IEEEkeywords}Multi Armed Bandits, Early-Exit, Natural Language Processing, Image Classification
% \end{IEEEkeywords}

\input{SPLITEE/chapters/abstract}

\input{SPLITEE/chapters/introduction}

\label{sec:intro}

\input{SPLITEE/chapters/related_works}

\input{SPLITEE/chapters/Problem_setup}

\input{SPLITEE/chapters/Algorithm}

\input{SPLITEE/chapters/Experiments}

\input{SPLITEE/chapters/Conclusion}

\bibliographystyle{IEEEtran}
\bibliography{mab,header}
\clearpage
\appendices
\section{More dataset description}	
\subsection{Feature selection}
The selected features from the MIMIC III and eICU datasets are shown in Table \ref{The selected features from the MIMIC III and eICU datasets } (due to limited space, for each category we show a maximum of 10 items).
\begin{table}[htbp]
 	\centering
        \small
 	\setlength{\tabcolsep}{2mm}
 	\caption{Examples of different categories of data in two EHR datasets. Due to space limitations only 10 entries per category are listed.}\label{The selected features from the MIMIC III and eICU datasets }
 	\begin{tabular}{ll}
 		\hline
 		\multicolumn{1}{c}{MIMIC III} &\multicolumn{1}{c}{eICU}
   \\ \hline
%demographic
\textbf{\tabincell{c}{\textit{Demographic}}}	&	\textbf{\tabincell{c}{\textit{Demographic}}} \\
 Age &gender \\
 Gender &age \\ 
Ethnicity  &ethnicity \\
 Marital Status &apacheadmissiondx\\
Insurance Type &admissionheight \\
Language &hospitaladmitsource \\
Religion &admissionweight \\
Admission Type &unitadmitsource \\
Admission Source &dischargeweight \\
DIAGNOSIS& unitdischargestatus\\
...& ...\\
%体征
\textbf{\textit{Vital signs}}&\textbf{\textit{Vital signs}} \\
SpO2&temperature        \\
Venous Pressure&  sao2      \\
Heart Rate&   heartrate     \\
DEEP BREATH&    systemicsystolic    \\
BIpap FIO2&  systemicdiastolic      \\
Skin Temperature&  systemicmean      \\
Daily Weight &  noninvasivesystolic      \\
real urine output&  noninvasivediastolic      \\
Cardiac output&  noninvasivemean      \\
FiO2&  paop      \\ 
...&   ...     \\

%lab test
\textbf{\textit{Lab test item}} & \textbf{\textit{Lab test item}}       \\
Red Blood Cell Count &Reticulocyte Count \\
	Iron &Red Blood Cells \\
Ferritin &Iron Binding Capacity \\
Hematocrit &Haptoglobin \\
Folate &Platelet Count \\
Haptoglobin &D-Dimer \\
Vitamin B12 &WBC  \\
Lactate Dehydrogenase &Bilirubin, Total, Ascites \\
Bilirubin Direct &glucose \\
Mean Corpuscular Volume  &Bilirubin \\
...  &... \\
%medication
\textbf{\textit{Medication}}  &\textbf{\textit{Medication}} \\
Vitamin B &VITAMINS/MINERALS \\
Folic Acid &WARFARIN SODIUM \\
Epoetin Alfa &Prednisone \\
Rituximab &SPIRONOLACTONE  \\
Warfarin &Azathioprine \\
Magnesium Oxide &NSAIDs \\
Docusate Sodium &Cyclophosphamide \\
Bisacodyl &Methotrexate \\
Acetaminophen&Hydroxyurea \\
Furosemide&Romiplostim \\
...&... \\
\textbf{\textit{ICD-9}} & \textbf{\textit{ICD-9}}\\
01166&414.00 \\
01170&491.20 \\
01171&428.0 \\
01172&427.31 \\
01173&585.9 \\
01174&414.00 \\
01175&428.0 \\
01176&585.9 \\
01181&427.31 \\
01182&518.82 \\
...&... \\

 \hline
 	\end{tabular}
 \end{table}

\subsection{Missing rate}
\begin{figure*}[htbp]
    \centering
    \includegraphics[width=6in]{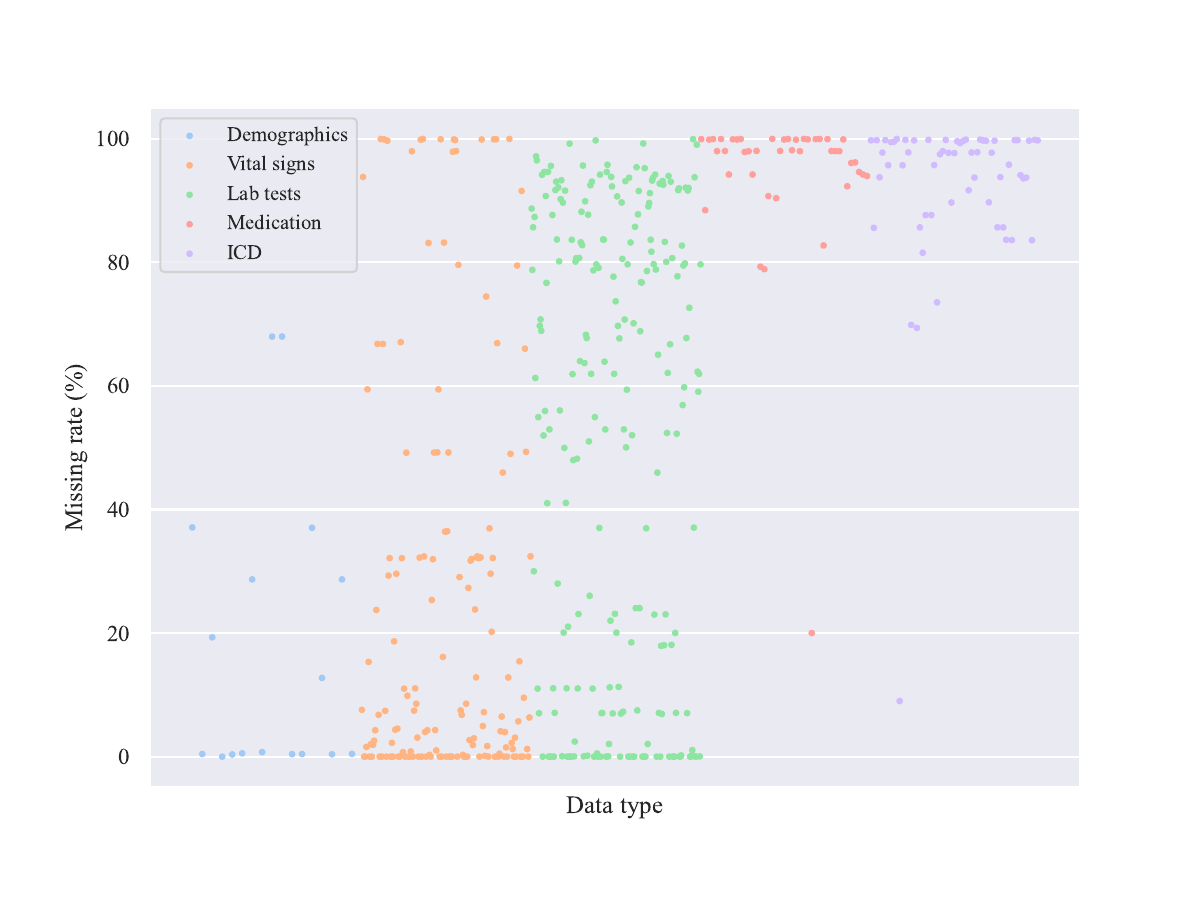}
    \caption{Visualization of missing data in MIMIC-III. The different coloured scatters represent that the feature comes from different data (i.e. Demographics, Vital, Lab, Medication, ICD)}
    \label{Missing rate MIMIC III}
\end{figure*}

\begin{figure*}[htbp]
    \centering
    \includegraphics[width=5in]{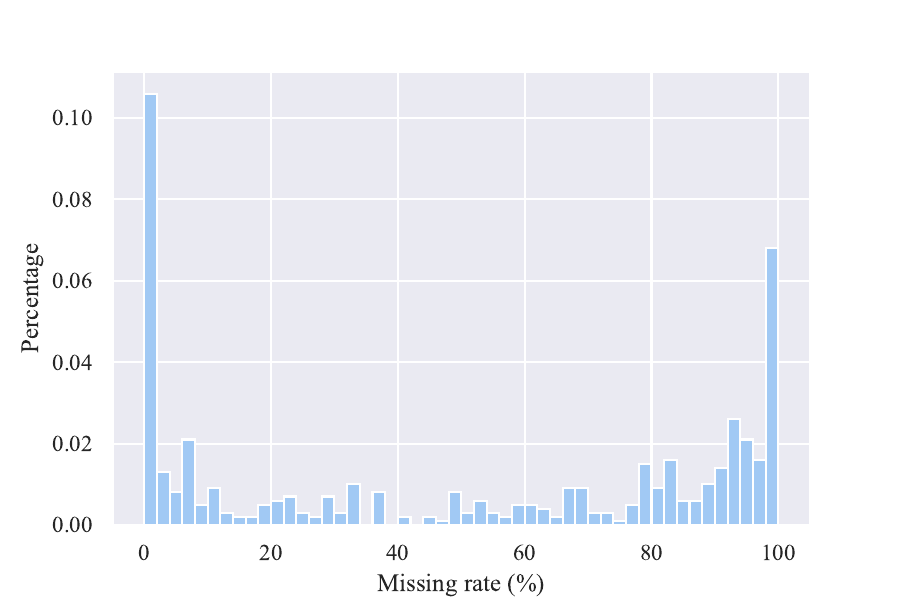}
    \caption{The histogram of the missing rate of MIMIC III dataset.}
    \label{hist MIMIC III}
\end{figure*}

\begin{figure*}[htbp]
    \centering
    \includegraphics[width=6in]{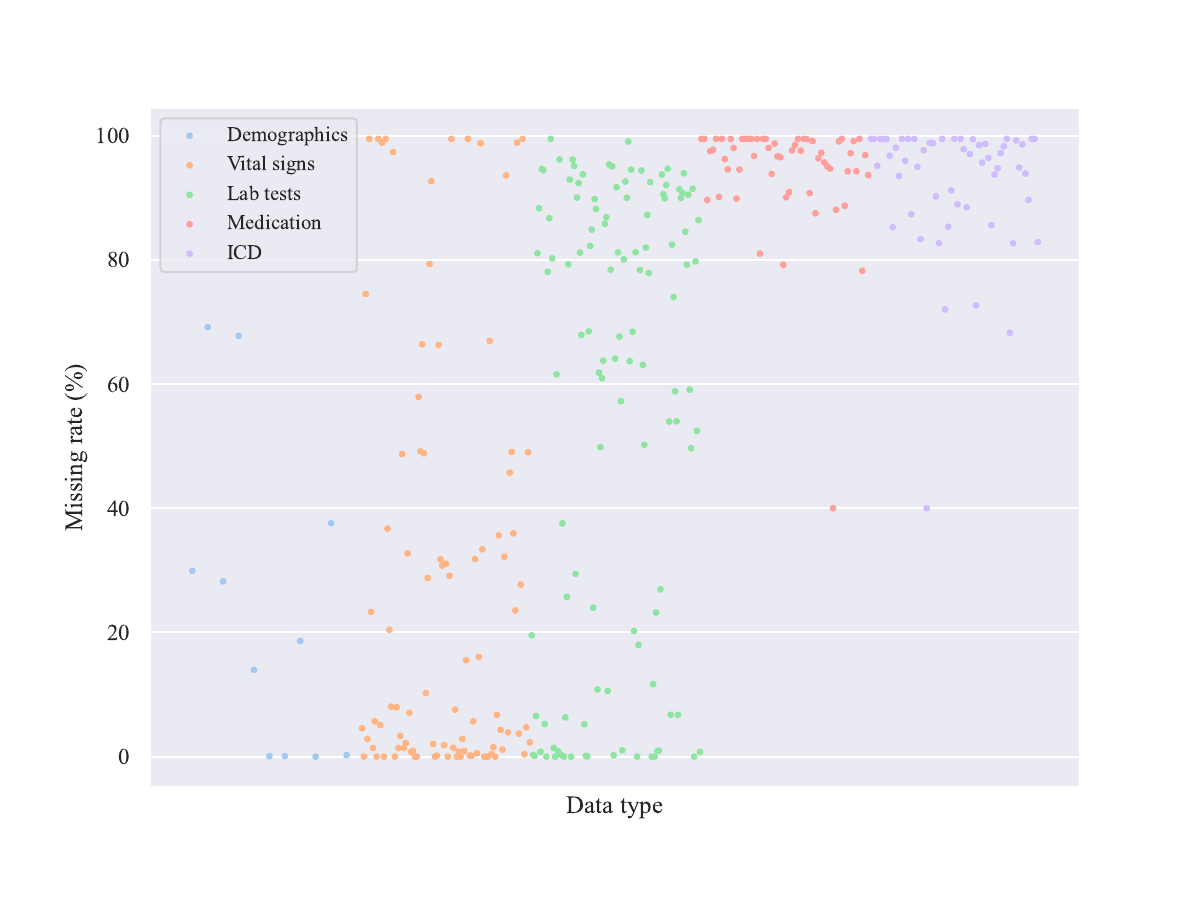}
    \caption{Visualization of missing data in eICU.}
    \label{Missing rate of eICU}
\end{figure*}

\begin{figure*}[htbp]
    \centering
    \includegraphics[width=5in]{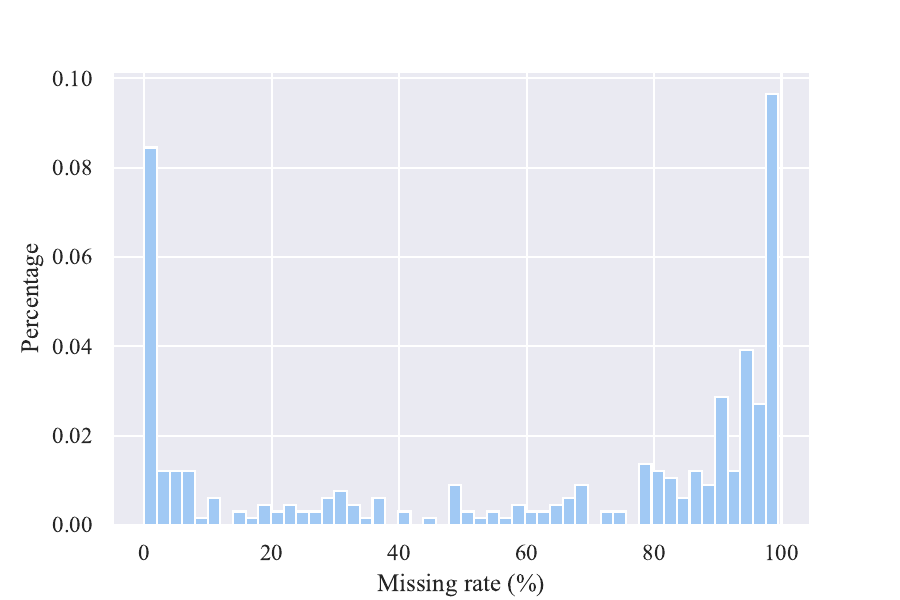}
    \caption{The histogram of the missing rate of eICU dataset.}
    \label{hist eICU}
\end{figure*}

The missing rate  for all the features that are finally involved (after preprocessing) with the histogram are shown in Fig. \ref{Missing rate MIMIC III} and Fig. \ref{hist MIMIC III} for MIMIC III dataset, and Fig. \ref{Missing rate of eICU} and Fig. \ref{hist eICU} for eICU dataset.  The statistical information on the missing rate in the MIMIC III and eICU dataset is given in the main paper.


% We give statistical information on the missing rate in the MIMIC III and eICU dataset in Tables \ref{statistical information on the missing data}.
% \begin{table}[htbp]
%  	\centering
%         \small
%  	\setlength{\tabcolsep}{1mm}
%  	\caption{\textcolor{black}{The statistical information on the missing data in the MIMIC III and eICU dataset. }}\label{statistical information on the missing data}
%  	\begin{tabular}{ccccc}
%  		\hline
   
%  		&\multicolumn{2}{c}{MIMIC III} &\multicolumn{2}{c}{eICU}
%    \\ 
%    Feature category
%    &\tabincell{c}{Feature\\number} & \tabincell{c}{Mean  missing\\ rate ($\%$)}   &\tabincell{c}{Feature\\number} & \tabincell{c}{Mean  missing \\ rate ($\%$)}\\ \hline
% Demographics &   17 & 17.84&11&20.61 \\
% Vital signs & 153 & 25.48&93&27.49 \\
% Lab tests & 228 & 51.95&116&50.96 \\
% Medication & 43 & 94.50 &58&96.91\\
% ICD & 59 & 92.59&55&93.64 \\
% All &500 & 51.14 &333 & 58.46 \\
%  \hline
%  	\end{tabular}
%  \end{table}

\section{More experimental results}

\begin{figure}[h]
\centering
\subfigure[use case1]{
\includegraphics[width=10cm]{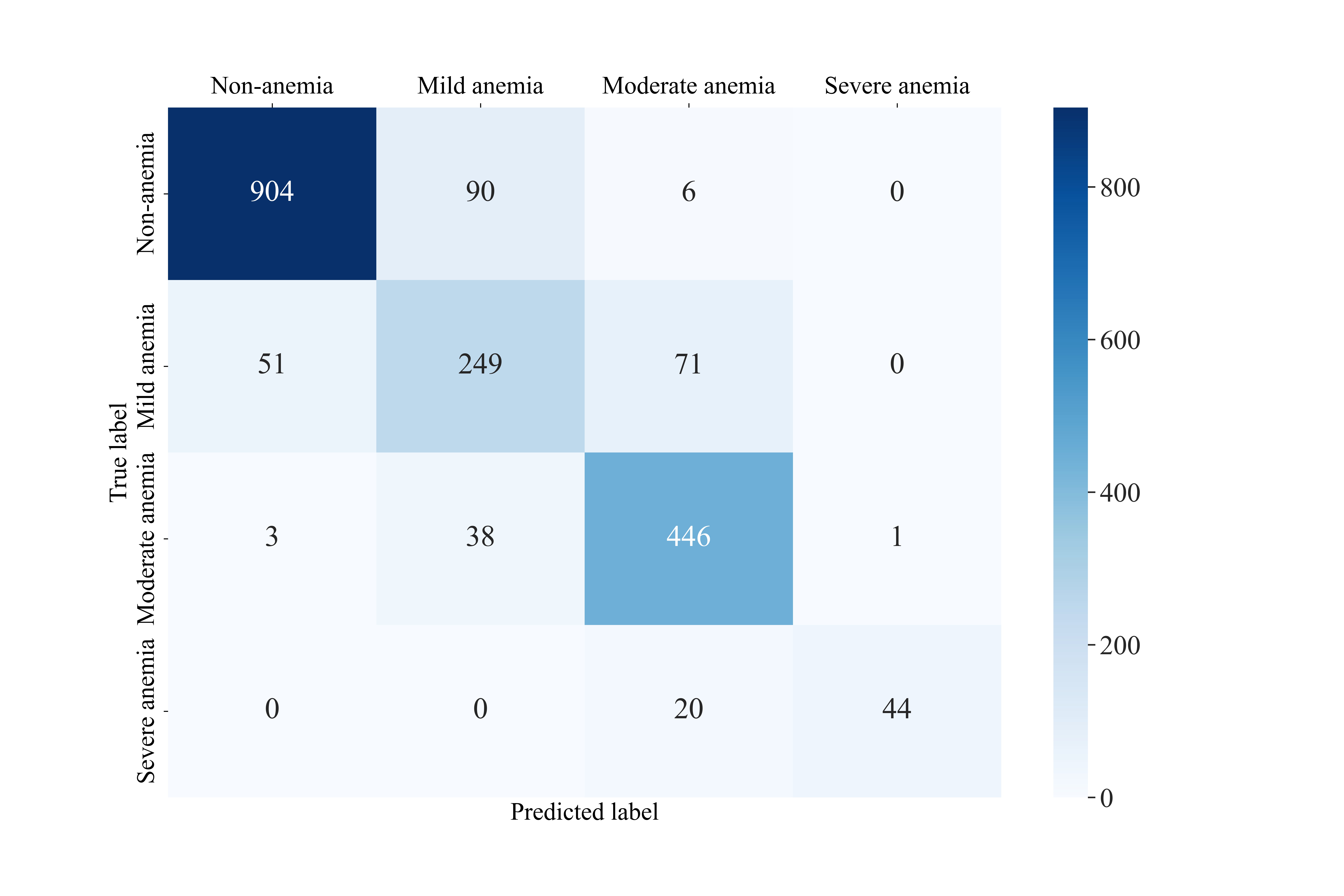}
%\caption{fig1}
}\hspace{-7mm}
\subfigure[use case2]{
\includegraphics[width=10cm]{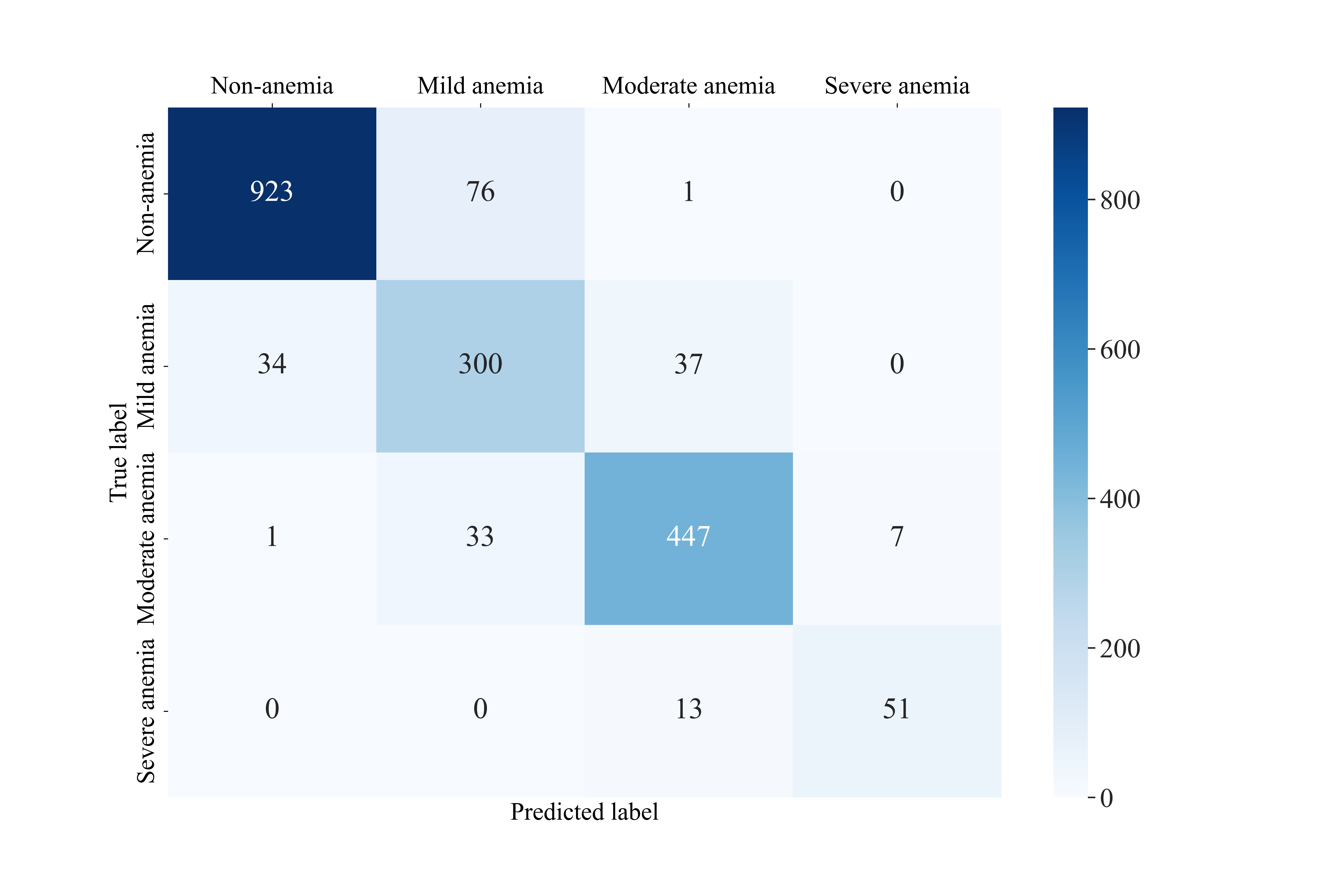}
%\caption{fig1}
}\hspace{-7mm}

\centering
\caption{The confusion matrix of the HgbNet in the MIMIC III dataset.}
\label{The confusion matrix of the ND-ATLSTM in the MIMIC III dataset.}
\end{figure}

\begin{figure}[h]
\centering
\subfigure[use case1]{
\includegraphics[width=10cm]{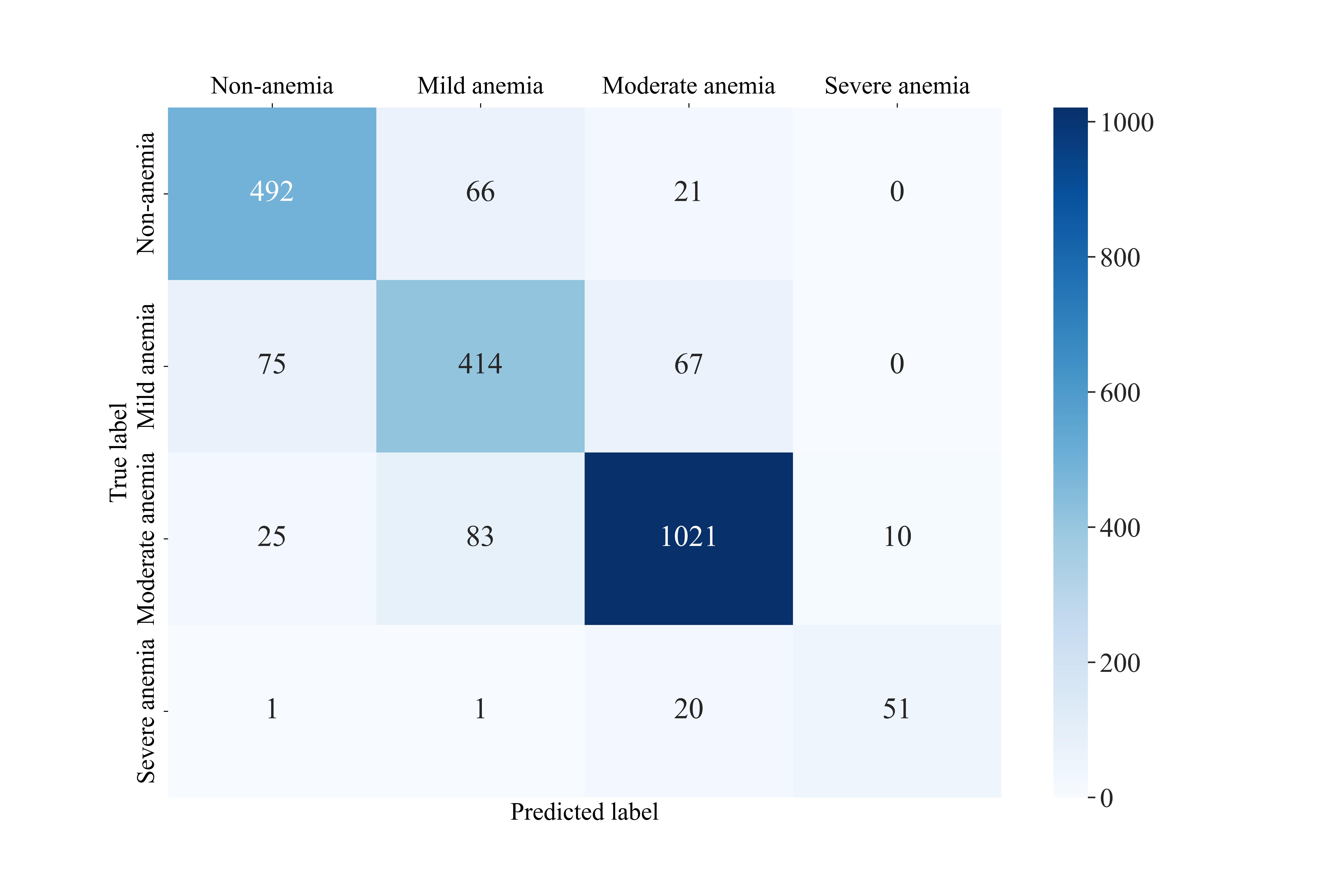}
%\caption{fig1}
}\hspace{-7mm}
\subfigure[use case2]{
\includegraphics[width=10cm]{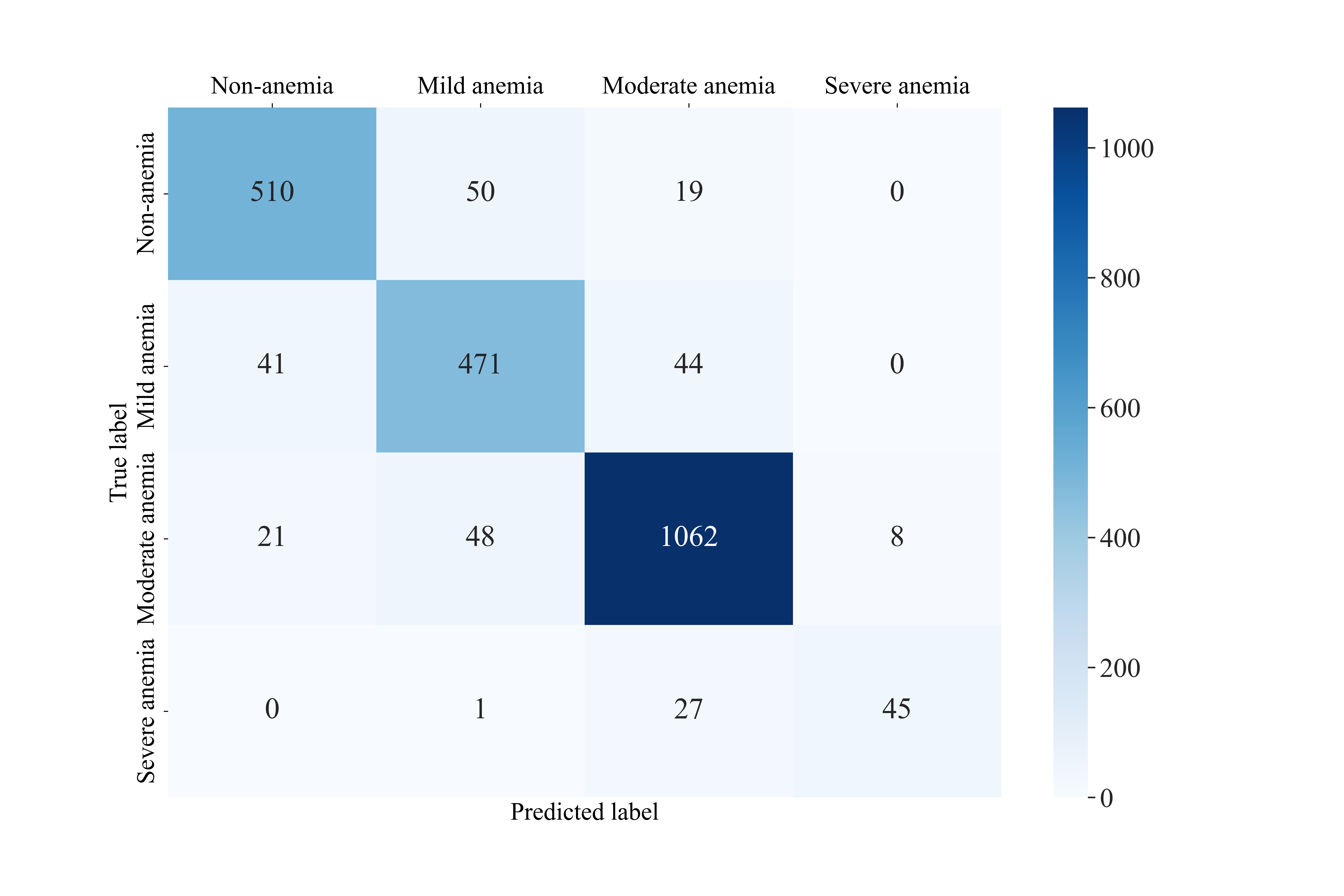}
%\caption{fig1}
}\hspace{-7mm}

\centering
\caption{The confusion matrix of the HgbNet in the eICU dataset.}
\label{The confusion matrix of the ND-ATLSTM in the eICU dataset.}
\end{figure}

To provide further insight into the results, we present the confusion matrix of HgbNet in the anemia degree prediction experiment in Fig. \ref{The confusion matrix of the ND-ATLSTM in the MIMIC III dataset.} and Fig. \ref{The confusion matrix of the ND-ATLSTM in the eICU dataset.}.

%mimic的混淆矩阵
%eicu的混淆矩阵

In Fig. \ref{The confusion matrix of the ND-ATLSTM in the MIMIC III dataset.}, the relationship between sample distribution across categories in anemia degree prediction results is evident.  As depicted  in  Fig. \ref{The confusion matrix of the ND-ATLSTM in the MIMIC III dataset.} (a), the test dataset contains 1000, 371, 488, and 64 samples for each label, respectively. The correctly diagnosed samples in these categories are 904, 249, 446, and 44, yielding accuracies of 90.4$\%$, 67.1$\%$, 91.4$\%$ and 68.8$\%$. This demonstrates that the sample size in each category significantly affects prediction outcomes.  Furthermore, we are concerned about potential biases in misdiagnoses, such as a healthy patient being diagnosed with severe anemia, which could have serious consequences. We define a serious misdiagnosis as a deviation of two or more labels from the true result (e.g., non-anemia to moderate anemia or mild anemia to severe anemia). In  use case 1, the number of serious misdiagnoses  for  the four labels is 6, 0, 3 and 0, with a serious misdiagnosis rate at 0.5$\%$, indicating a high  confidence of the model. In use case 2, the accuracy of each label is improved by 2.1$\%$, 20.5$\%$, 0.2$\%$ and 15.9$\%$. The number of serious misdiagnoses for each of the four labels is 0, 0, 1 and 0, demonstrating that serious misdiagnoses have been almost eradicated.  As shown  in Fig. \ref{The confusion matrix of the ND-ATLSTM in the eICU dataset.},  the serious misdiagnosis counts in the two cases are 21, 0, 25, 2 and 19, 0, 21, 1 (with the test dataset containing 579, 556, 1139, and 73 samples for each label) , and the serious misdiagnosis rates are 2.0$\%$ and 1.7$\%$, indicating a decreasing trend.

In summary,  the performance of the HgbNet  on both tasks and use cases has been  improved by providing certain test values at the moment $T+1$.

\end{document}

%% file: SPLITEE/chapters/abstract.tex
\begin{abstract}
Anemia is a prevalent  medical condition that typically requires  invasive blood tests for diagnosis and monitoring. Electronic health records (EHRs) have emerged as valuable data  sources  for numerous  medical studies.
EHR-based hemoglobin level/anemia degree prediction is non-invasive and rapid  but  still faces some challenges
due to the fact that EHR data is typically an irregular  multivariate  time series  containing a significant number of missing values and irregular time intervals. To  address these issues, we introduce  HgbNet, a machine learning-based prediction model that emulates clinicians' decision-making processes for hemoglobin level/anemia degree prediction.    The model incorporates a NanDense layer with a missing indicator to handle missing values and employs attention mechanisms to account for   both local irregularity and global irregularity.  We evaluate the proposed method using two real-world datasets across  two use cases. In our first use case, we predict  hemoglobin level/anemia degree at  moment  $T+1$ by utilizing records from moments prior to  $T+1$. In our second use case, we integrate   all  historical records with additional selected  test results at moment $T+1$ to predict hemoglobin level/anemia
degree  at the same moment, $T+1$.   HgbNet outperforms the best baseline results across all datasets and use cases. These findings demonstrate the feasibility of estimating hemoglobin levels and anemia degree from EHR data, positioning HgbNet as an effective non-invasive anemia diagnosis solution that could potentially enhance the quality of life for millions of affected individuals worldwide. To our knowledge, HgbNet is the first machine learning model leveraging EHR data for hemoglobin level/anemia degree prediction.
\end{abstract}
\begin{IEEEkeywords}
Anemia/hemoglobin prediction, EHR, attention, LSTM, missing values.
\end{IEEEkeywords}

%% file: SPLITEE/chapters/introduction.tex
\section{Introduction}
Anemia  is a widespread  blood disorder affecting  approximately 1.6 million people globally  \cite{mclean2009worldwide}. As reported by  \cite{wangzhiyi}, the worldwide  prevalence of anemia among all age groups in 2019 was  22.8$\%$ (95$\%$ CI: 22.6–23.1). Given its significant contribution to global disease, prompt diagnosis and treatment of anemia are crucial for enhancing the quality of human life.

Traditional anemia diagnosis involves measuring hemoglobin concentration through venous blood samples, a process that is invasive and requires a clinical or outpatient setting.  Moreover, this method can cause pain, localized infection, and generate medical waste. Consequently, researchers are investigating non-invasive approaches for hemoglobin/anemia estimation to alleviate the burden on clinical and laboratory services, particularly where frequent monitoring is essential.

 Electronic health record (EHR) systems  store comprehensive patient data like demographics and lab test results \cite{shickel2017deep}. 
 % are increasingly utilized in research for disease prediction \cite{si2021deep}, clinical decision-making \cite{schwartz2021clinician}, and medical concept extraction \cite{li2022neural}. 
EHR-based disease prediction offers a non-invasive, cost-effective approach and many machine learning models have been proposed to diagnose conditions such as clinical subtype, septic shock, in-hospital mortality, and phenotyping from EHR data \cite{baytas2017patient,tan2020data,zhang2019attain,lee2022multi,luo2020hitanet}. However, to our knowledge, 
% Clinically, EHR systems record patient history, enabling blood result trend monitoring, facilitating early anemia detection, and providing a warning system for patients at risk of acute or chronic anemia.  Utilizing predictive methods with machine learning could potentially reduce patient harm, improve safety, compliance, and enhance overall health outcomes through regular monitoring. 
no  models  specifically  predict hemoglobin level/anemia degree using EHR data. 

\begin{figure}[htbp]
	\centering
	\includegraphics[width=3.5in]{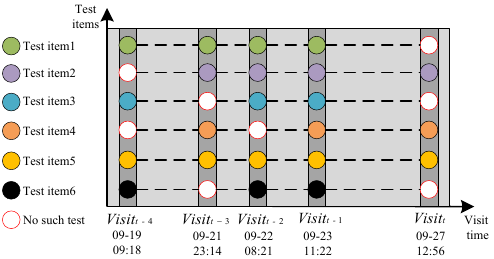}
	\caption{An example of a patient's partial EHR. \textcolor{black}{The patient had 5 visits from 09-19-2022 09:18 to 09-27-2022 12:56 (from $Visit_{t-4}$ to $Visit_{t}$) with varying time intervals between consecutive visits, illustrating the global irregularity (the irregularity of each visit). In each visit, the patient may have undergone some of the six test items, leading up to missing values  between consecutive results for each test.   This phenomenon exemplifies local irregularity (the irregularity of each test item)}.    In actual EHR data, time intervals can range from hours to months, and the number of test items can reach up to 1,000. Such irregular time series pose considerable challenges for disease prediction.}
	\label{An example of a patient partial EHR}
\end{figure}

This study aims to explore the technologies employed for this purpose. However, we are confronted with various challenges in developing machine-learning models that ingest EHR data. Specifically, for a patient, 1) the time interval between visits can vary considerably (from hours to months), referred to as global irregularity,  and 2) test items  may differ during each visit, referred to as local irregularity. Overall, these two issues lead to an irregular multi-variate time series exhibiting various missing values. These challenges are  illustrated in Fig. \ref{An example of a patient partial EHR}.

There have been various methods proposed to deal with such challenges.  For dealing with missing values in time series, numerous  imputation algorithms have been proposed, including   single imputation \cite{zhang2016missing,le2021sa} and multiple imputations \cite{van1999flexible,azur2011multiple,stekhoven2012missforest},  as well as  neural network-based methods  such as GAN and RNN\cite{luo2018multivariate,suo2019recurrent}. However, imputation-based techniques are known to distort the original data, thereby influencing any prediction/diagnostic result/judgment. 
% Moreover,  imputation introduces additional computational burdens when dealing with high-dimensional data.

In turn, for dealing with irregular time intervals in time series,  one can adopt interpolation and alignment techniques to regularize an irregular time series \cite{marlin2012unsupervised, lipton2016directly}; one can also adopt models that directly ingest irregular time series by using techniques such as a time-decay function \cite{baytas2017patient,zhu2017next,ruan2022real}, ordinary differential equations \cite{rubanova2019latent,lechner2020learning}, or time embedding modules \cite{yin2020identifying,lee2022multi,zhang2021merits}. However, such approaches have primarily been developed to deal with global regularity issues in \emph{lieu} of local irregularity.  Clinicians however diagnose a disease from EHR data by examining the target disease history, considering both global trends and local ones. 
% The more specific illustration can be found in Fig. \ref{An example of a patient partial EHR}.

To address the aforementioned limitations, this study introduces HgbNet,  a model designed to predict hemoglobin levels and anemia degrees using EHR data. HgbNet comprises   two primary  components: 1) Missing data processing: The model employs a NanDense layer with missing indicators to mitigate the impact  of missing values. Rather than imputing the original data, the NanDense layer  adaptively processes  missing values by adjusting neuron activations in the first embedding layer  \cite{kia2022promissing}. This approach  adds no supplementary parameters or computational burden  while maintaining competitive performance.   Furthermore, in the same way, as clinicians only refer to observed test values, the missing indicator serves to distinguish between observed and missing data, enabling the model to allocate varying attention levels accordingly. 2) \textcolor{black}{Global vs local irregularity processing: Our model employs three attention modules to mimic a clinician's decision process. Label-level and feature-level attentions address two types of irregularities, while general attention calculates the interaction of individual records.}

We have assessed the efficacy of our approach in comparison to various baselines by conducting a series of experiments on two real-world EHR datasets: MIMIC III and eICU. Two use cases are also examined:  1) Predicting a patient's  hemoglobin level/anemia degree at moment/visit $T+1$  using data prior to moment/visit $T+1$ (i.e. a patient's visit from moment 1 to $T$). From a clinical perspective, this scenario  is applicable to patients who are unavailable for immediate testing (e.g., emergency patients) and rely solely on historical data. 2) Predicting  a patient's  hemoglobin level/anemia degree at moment/visit $T+1$  by incorporating additional selected test values for moment/visit $T+1$, along with all previous  records.  This case caters to patients without urgent requirements who can undergo simple tests upon a current visit to the healthcare provider. The objective is to determine if incorporating non-invasive, real-time measurements (e.g., pulse, blood pressure) enhances prediction accuracy.  Experimental results demonstrate the feasibility of estimating hemoglobin levels and anemia degrees from EHR data in these two use cases, with the proposed model outperforming state-of-the-art methods. 

The contributions of this study are summarized as follows:

$\bullet$ HgbNet is to the best of our knowledge the first model introduced for predicting hemoglobin levels and anemia degrees from EHR data, addressing challenges such as irregular time intervals and missing values. 

$\bullet$ The proposed method is evaluated against other state-of-the-art approaches on two real-world EHR datasets. The experiments cover two use cases, encompassing common application scenarios.

$\bullet$  Experimental results illustrate the feasibility of predicting hemoglobin levels and anemia degrees from EHR data with the proposed method in comparison to various baselines.

The remainder of this article is organized as follows: Section II provides a review of related work in the field. Section III introduces our approach -- HgbNet -- to predict hemoglobin level and anemia degree from EHR data. Section IV outlines the experimental setup and Section V presents  the results. Lastly, Section VI concludes the study, proposing also potential directions for future research.

%% file: SPLITEE/chapters/related_works.tex
\section{Related works}
 %这个重新写，就是相当于把intro里面的写仔细一些
We begin by offering an overview of  the related work addressing the key challenges in processing EHR data.
\subsection{Methods addressing irregular time intervals in time series}
The existing literature on handling irregular time intervals in time series can be broadly categorized into three groups.

1) \textit{Alignment and interpolation approaches}. Recurrent neural network (RNN) and long short-term memory (LSTM) are popular for time series analysis but struggle with irregular clinical time series due to  these methods have been primarily designed to deal with time series with fixed time intervals \cite{sherstinsky2020fundamentals}.    A common solution involves interpolating and aligning data time intervals \cite{marlin2012unsupervised,lipton2016directly}. However, this approach introduces considerable uncertainty/noise in the presence of highly irregular time series  and requires manual selection of time intervals, potentially compromising performance.

2) \textit{Modeling irregular time series directly}.  \textcolor{black}{A more effective strategy is to use irregular time intervals to adjust the information transfer  of the sequence model.  For instance, a monotonically non-increasing function can be used to transform time intervals into the weight of long-term memory in LSTM \cite{baytas2017patient}. Similarly, three  LSTM structures   with time gates are proposed \cite{zhu2017next}.  Such modifications  are also performed on the gated recurrent unit (GRU) \cite{ruan2022real}. Alternatively, ordinary differential equations (ODE) can use continuous time dynamics to represent hidden states and are combined to build ODE-RNN \cite{rubanova2019latent} and ODE-LSTM \cite{lechner2020learning}, which improve performance in irregular time series tasks.  However, these sequential models may struggle with processing long-time series.}

3) \textit{Attention mechanism-based methods}. Recent attention mechanisms have been employed to handle irregular time series efficiently and enable parallel processing of long series   \cite{vaswani2017attention}. For example, the time-attention based on T-LSTM, has been proposed to reassign time weights of all preceding moments and achieve accurate disease progression modeling \cite{zhang2019attain}.  Similar studies can also be found in \cite{tan2020data,cheng2022multimodal}. 
Another common approach is to  combine local attention of the hidden layer  with global attention transformed by irregular time intervals to account for input data features 
\cite{luo2020hitanet,liu2022catnet,cheng2021ggatb}. Additionally, some research  replaces   sequential models with transformers, embedding irregular time intervals into the model inputs to simplify  the process \cite{lee2022multi}. Other attention mechanism-based methods  are reported in \cite{xia2022mulhita,zeng2023time,an2022tertian}.
% \end{enumerate}

\textcolor{black}{Despite the attention-based approach demonstrating superior outcomes, the existing methodology solely focuses on global irregularity, neglecting local irregularity. Consequently, this can result in the loss of intricate details in EHR. To address this issue, our study incorporates three distinct attention mechanisms: local attention, feature-level attention, and label-level attention. These mechanisms cater to individual record interactions, local irregularities, and global irregularities, respectively.}

\subsection{Methods addressing missing values in time series}
The related work dealing with  time series with missing values can be summarized to two categories.

1) \textit{Imputation-based methods}. Imputation-based methods are common for handling missing data.  Single imputation methods, like mean/mode calculation or constant selection, are simple but susceptible to noise  \cite{zhang2016missing, le2021sa}.  Multiple imputation methods estimate missing values by modeling observed values and missing patterns.  Representative multiple imputation methods include MICE \cite{azur2011multiple}, KNN \cite{zhang2012nearest}, MissForest \cite{stekhoven2012missforest}, GAN \cite{luo2018multivariate}, and RNN \cite{suo2019recurrent}.  Despite achieving more accurate results through complex iterative and computational processes, multiple imputation methods impose a greater computational burden, especially for high-dimensional data, and perform worse with a large proportion of missing values.

2) \textit{Non-imputation-based methods}. Missing indicators have been applied to handle missing values \cite{lee2022multi,yin2020identifying,nijman2022missing}. This approach involves using a binary variable for each data point to indicate if the value is missing while replacing all missing values with a constant. This method is simple but its effectiveness highly depends on the selection of the constants.  Less common approaches for handling missing values include likelihood-based methods \cite{shin2022evaluation} and  submodels methods \cite{stempfle2022sharing}.

\textcolor{black}{The performance of imputation-based methods are limited when dealing with high-dimensional data, with most alternative approaches necessitating manually set parameters. Inspired  by \cite{kia2022promissing}, we employ NanDense, an adaptive method for handling missing values without extra parameters, allowing neurons to auto-adjust their response. Furthermore, we incorporate missing indicators  to integrate prior knowledge, enabling the model to focus attention on missing values distinct from observed data.}

%% file: SPLITEE/chapters/Problem_setup.tex
\section{Hemoglobin level/anemia degree prediction using HgbNet}We are now ready to introduce our prediction problem along with our approach.

%请注意这里的time interval是连续两个visit之间的差值，而不是每一个与最后一个差值,因此在后面要区别说明一下
We model the EHR data associated with a single patient using a multivariate time-series $X=\{x_1,x_2,...,x_T\}$, comprising up to $T$ patient visits. We also model the data associated with a patient visit using a multi-dimensional vector $x_t=\{x_t^1,x_t^2,..., x_t^{K}\}$ comprising $K$ test items \footnote{These test items could be associated with physiological features and lab test results}.  The time interval between two consecutive visits is recorded in a $T$-dimensional vector given by $\Delta=\{\Delta_1,\Delta_2,...,\Delta_ T\}$, $\Delta_1=0$.
Since  a patient's visit depends on clinical needs, $\Delta_t$  can vary  from  hours to months. We also note that due to the fact that a patient may potentially need different test items during each visit, the multi-variate time-series $X$ may contain numerous missing values. Fig. \ref{An example of a patient partial EHR} illustrates an example segment of the EHR record for a single patient. 

This study considers two use cases. In use case 1, all of a patient's historical records $X$ are utilized to predict the hemoglobin level/anemia degree at the next moment ($T+1$) as follows: 
% In use case 2, non-invasive and easily accessible test results at moment $x_{T+1}$ (details of $x_{T+1}$ are described in Section \ref{HAHA}) are combined with all historical records $X$ to predict the hemoglobin level/anemia degree in the next moment.  The mathematical descriptions of the two use cases are expressed as follows:
\begin{equation}
\begin{gathered}
    \hat{y}_{h{(T+1)}}=h_{\theta1}((x_1,x_2,...,x_T)),\\
    \hat{y}_{a{(T+1)}}=h_{\theta2}((x_1,x_2,...,x_T)),
\end{gathered}
\end{equation}
where $h_{\theta_1} (\cdot)$ denotes a predictor parameterized by a series of parameters $\theta_1$ that delivers the patient's hemoglobin level at time $T+1$ given the patient's record up to time $T$ whereas $h_{\theta_2} (\cdot)$ denotes the predictor parameterized  by $\theta_2$ for anemia degree prediction. Note that prediction of the hemoglobin level is a regression problem because the hemoglobin level typically ranges from 5-20g/dL whereas prediction of the anemia degree is a multi-class classification problem comprising the classes Non-anemia, Mild anemia, Moderate anemia and Severe anemia. $\hat{y}_h (T+1)$ and $\hat{y}_a (T+1)$ denotes the hemoglobin level and anemia degree at time $T+1$, respectively.  

In use case 2, non-invasive and easily accessible test results $x_{T+1}$ at moment ${T+1}$ (details of $x_{T+1}$ are described in Section \ref{HAHA}) are combined with all historical records $X$ to predict the hemoglobin level/anemia degree at time $T+1$ as follows:
\begin{equation}
\begin{gathered}
\hat{y}_{h{(T+1)}}=h_{\theta_1^{\prime}}((x_1,x_2,...,x_{T+1})),\\
\hat{y}_{a{(T+1)}}=h_{\theta_2^{\prime}}((x_1,x_2,...,x_{T+1})),\\
    x_{T+1}=\{x_{T+1}^1,x_{T+1}^2,..., x_{T+1}^{L}\},
\end{gathered}
\end{equation}
where $h_{\theta_1^{\prime}} (\cdot)$ and $h_{\theta_2^{\prime}} (\cdot)$  denote the predictor parameterized by  $\theta_1^{\prime}$ and $\theta_2^{\prime}$, predicting the hemoglobin level and   anemia degree, respectively. $L$ refers to the number of test results involved at the moment $x_{T+1}$ in use case 2. We next describe the structure of the predictors i.e. HgbNet.
\subsection{HgbNet}\label{HgbNet}
The proposed  HgbNet's architecture is illustrated in Fig. \ref{The structure of the ND-ATLSTM model.}. It comprises a series of modules that are described further in the sequel: 1/ the input representation module; 2/ a NanDense layer; 3/ a time embedding module; 4/ an LSTM-M module; 5/  local, feature level and label level attention mechanisms; and 6/ the regressor/classifier providing a prediction of the hemoglobin level/anemia degree.
\begin{figure*}[h]
\label{The structure of the ND-ATLSTM model.}
	\centering
	\includegraphics[width=6in]{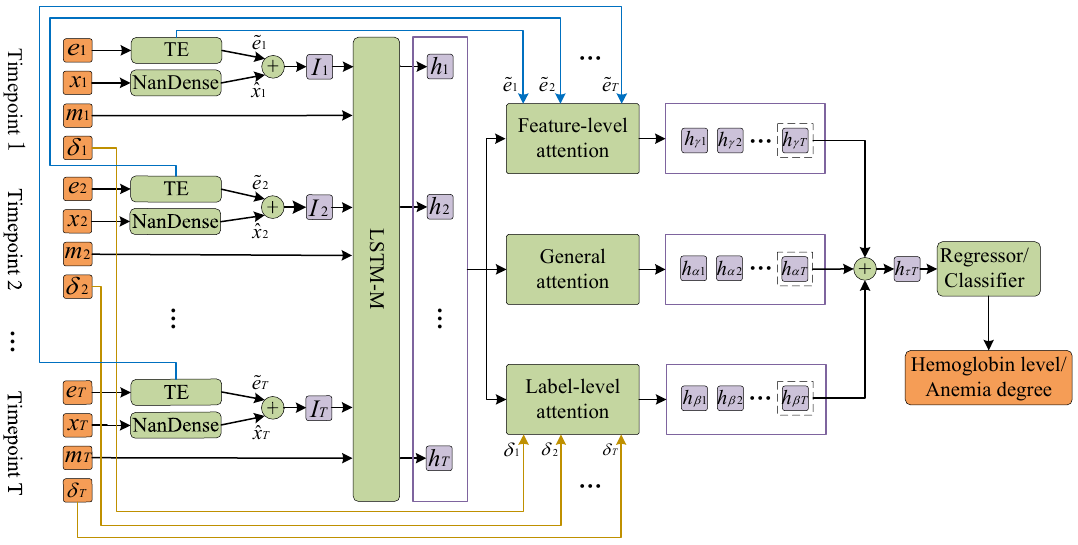}
	\caption{The structure of the proposed HgbNet. At each time step $t$, the HgbNet input comprises four components: the original EHR data $x_t$, the feature-specific time interval matrix $e_t$, the missing indicator $m_t$, and the label time interval matrix $\delta_t$. The time embedding (TE) layer and NanDense layers process $e_t$ and $x_t$, respectively, before being input to the LSTM-M network alongside $m_t$ to generate the hidden representation $h_t$. Subsequently, $\{h_1,h_2,...,h_T\}$ interacts with itself, $\{\hat{e}_1,\hat{e}_2,...,\hat{e}_T\}$, and $\{\delta_1,\delta_2,...,\delta_T\}$ to compute three attention types, accounting for each record's interaction, local irregularity, and global irregularity. Finally, the fused hidden representation $h_{\tau T}$, derived from the three attention results, is employed for downstream tasks to predict hemoglobin level and anemia degree at time step $T+1$.}

 % The patient's EHR is fed into the HbgNet sequentially according to the visit time.  For  moment $t$, the input is represented by four components: the original EHR data $x_t$, the time interval matrix for each feature $e_t$, the missing indicator $m_t$ and the time interval matrix for label $\sigma_t$. $e_t$ and  $x_t$ are processed by the TE and NanDense layers respectively, and then fed into the LSTM-M network together with $m_t$ to obtain the hidden representation $h_t$.  Next, $h_1,h_2,...,h_T$ interacts with itself, $\{\hat{e}_1,\hat{e}_2,...,\hat{e}_T\}$ and $\{\sigma_1,\sigma_2,...,\sigma_T\}$ to calculate  three types of attention, considering the interaction of each record, the local irregularity and the global irregularity, respectively. Finally, the hidden representation $h_{\tau T}$, fused from the three types of attention results, is used to perform the  downstream task to get the predicted hemoglobin level/ anemia degree of moment $T+1$.}
	\label{The structure of the ND-ATLSTM model.}
\end{figure*}

\subsubsection{Input representation module}
For each patient, EHR data are sequentially fed into the model according to the visit time. At  moment $t$, four inputs are extracted from the EHR 
record:

    $\bullet$ $x_t\in \mathbb{R}^{1 \times K}$ represents the original EHR data, which may contain numerous missing values. These missing values should not be set to 0, as they signify unmeasured items  rather than a measured value of 0.  The NanDense layer processes these missing values, transforming them into acceptable inputs for the model while preserving potential information (the NanDense layer is described in the sequel).
    
    $\bullet$ $e_t\in \mathbb{R}^{1 \times K}$  represents a vector of feature time intervals, capturing local irregularity. In particular, if $x_t^{K}$ is observed, $e_t^{K}$ denotes the time difference between $t$ and $T$; otherwise, $e_t^{K}$ represents the time difference between the last moment $t_l$ when $x_{t_l}^{K}$ is observed and $T$.
    
    $\bullet$ $m_t\in \mathbb{R}^{1 \times K}$  represents a vector that indicates whether  particular test items in time $t$ are missing. Specifically, if a value in $x_t$ is missing, the corresponding entry in $m_t$ is assigned 0; otherwise, it is assigned 1. Incorporating missing indicators imparts a priori knowledge to the model, enabling it to differentiate between observed and missing values. 
    
    $\bullet$ $\delta_t \in R$ represents the time interval between $t$  and the latest visit $T$, accounting for global irregularities.
    %请注意 x=[x1,x2,xT]这里面最后一个用T，然后说每一个用xt,后面统一一起改

\subsubsection{NanDense layer}
The EHR record $x_t$ is initially processed by the NanDense layer. Data containing 'Not a Number' (NaN) values are unsuitable for direct input into the LSTM-M model.  Furthermore, imputation or deletion-based methods often present numerous challenges. The NanDense layer serves as an optimized Dense layer designed to address missing value issues, particularly in cases with a high proportion of missing data. Rather than interpolating the data, the NanDense layer mitigates the impact of incomplete datasets by adjusting the internal structure of the Dense layer \cite{kia2022promissing}.

 As previously mentioned, $x_t$ is a vector consisting of $K$ features. Due to the  presence of a substantial number of missing values in $x_t$, it is divided into two parts: the available data $x_t^a$ containing $q$ features (or test items) and the missing data $x_t^m$ containing $r$ missing values  ($K=q+r$). When $x_t$ is input into a neural network model with a Dense layer, the activation of the $v$th ($v$ is determined as needed) neuron in the Dense layer can be expressed as:
\begin{equation}\label{activation}
f^{(v)}=\sum_{x_{ti} \in {x_t}^a} x_{ti} w_i^{(v)}+\sum_{x_{tj}\in {x_t}^m} x_{tj} w_j^{(v)}+b^{(v)},
\end{equation}
where $f^{(v)}$ denotes  the neuron's activation, $w$ represents  the neuron's weight and $b$ indicates  the bias.  Eq. \ref{activation} can not be implemented where $x_t$ contains various missing values. Therefore, to address this issue, the equation's term that involves missing data (second term in Eq. \ref{activation}) is removed, and the bias is adjusted to neutralize the influence. The modified  activation is expressed  as follows: 
% \begin{equation}\label{new}
%     f^{(v)}=\sum_{x_{ti} \in {x_t}^a} x_{ti} w_i^{(v)}+\frac{q b^{(v)}}{K}.
% \end{equation}
% As evident from Eq. \ref{new}, two actions are performed on the original activation:  1)excluding the missing data from the activation by eliminating the corresponding weight calculation, and 2) attenuating the effects of missing data by adjusting the bias with a factor of $q/K$.

% Eq. \ref{new} may encounter  challenges when  handling   high-dimensional data with a high percentage of missing values ($K \rightarrow \infty$ and $r \gg q$). In this scenario,  $\frac{q b^{(v)}}{K}$  converges to 0, potentially resulting in an excessively small neuron output and adversely affecting the model's performance. To address  this issue, a compensatory weight $w_c$ is introduced and Eq. \ref{new} is modified  as:
\begin{equation}\label{haha}
    f^{(v)}=\sum_{x_{ti} \in {x_t}^a} x_{ti} w_{ti}^{(v)}+\frac{q b^{(v)}+r w_c^{(v)}}{K},
\end{equation}
where $w_c$ is the compensation factor to avoid insufficient neuronal output when handling high-dimension data ($K \rightarrow \infty$ and $r \gg q$). Note that $w_c$ is learned during the training process. Upon  processing in the NanDense layer, the vector $x_t\in \mathbb{R}^{1 \times K}$ is embedded  onto a vector $\hat{x}_t\in \mathbb{R}^{1 \times K}$. In summary, two actions are performed on the original activation:  1) excluding the missing data from the activation by eliminating the corresponding weight calculation, and 2) attenuating the effects of missing data by adjusting the bias.

The benefits of the NanDense layer can be summarized as follows: 1) the absence of additional parameters or models ensures computational efficiency (\textcolor{black}{only the first Dense layer is replaced with NanDense layer}), particularly for high-dimensional data, and 2) all parameters are derived from the training and learning phases, enabling the data-driven approach to fully optimize the parameters, especially in the context of large-scale data sets.
\subsubsection{Time embedding layer}
The time embedding (TE) layer is devised to modify the feature's time interval vector $e_t\in \mathbb{R}^{1 \times K}$, onto another vector $\tilde{e}_t \in \mathbb{R}^{1 \times K}$ -- with the goal to prioritize more recent test results -- that will be later embedded onto the  modified record $\hat{x}_t \in \mathbb{R}^{1 \times K}$. This is achieved as follows:

\begin{equation}
\begin{gathered}
\textcolor{black}{
f_t=\left(W_{e1} \odot \frac{e_t}{\delta_{max}} + b_{e1}\right),}\\
\textcolor{black}{\widetilde{e_t}= W_{e2} \odot \left(v - tanh\left(f_t \odot f_t\right)\right) + b_{e2},}
\end{gathered}
\end{equation}
\textcolor{black}{where $\odot$ represents  element-wise multiplication, $W_{e1}, b_{e1}, W_{e2}, b_{e2}\in \mathbb{R}^{1 \times K}$ represent  the learnable parameters.   $tanh$ refers to the hyperbolic tangent activation function \cite{APICELLA202114}. All  elements of $v\in \mathbb{R}^{1 \times K}$ are 1 and $\delta_{max}$ is the maximum value of $\delta_t$}.
Test results from a patient's distant past have limited relevance to their current diagnosis, and the TE layer effectively diminishes their impact. We then  add the  output of the NanDense layer, denoted as $\hat{x}_t$, to $\widetilde{e}_t \in \mathbb{R}^{1 \times K}$ to obtain a data representation  $I_t = \hat{x}_t + \widetilde{e}_t$
incorporating local irregularity-awareness.
\subsubsection{LSTM-M}
We design an LSTM-M network -- ingesting $I_t \in \{I_1, I_2,..., I_T\}$ and $m_t \in \{m_1,m_2,...,m_T\}$ and delivering 
$h_t$ -- to learn the long and short-term dependence in  the EHR data at different time points. 
% In a traditional LSTM, the time step (the first dimension of the input) for each  input moment can be set manually. For simplicity, we set the time step to 1 in the following equations. 
In particular, the LSTM-M network is defined  as:
\begin{equation}
\begin{gathered}
f_t=\sigma\left(W_f I_t^\top+U_f {h}_{t-1}+V_f m_t^\top+b_f\right),\\
i_t=\sigma\left(W_i I_t^\top+U_i {h}_{t-1}+V_i m_t^\top+b_i\right), \\
o_t=\sigma\left(W_o I_t^\top+U_o {h}_{t-1}+V_o m_t^\top+b_o\right),\\
c_t=f_t \odot c_{t-1}+i_t \odot tanh\left(W_c I_t^\top+U_c {h}_{t-1}+V_c m_t^\top+b_c\right),\\
h_t=o_t \odot tanh\left(c_t\right),
\end{gathered}
\end{equation}
 where $h_t \in \mathbb{R}^{H \times 1}$ is the LSTM-M network output at time step $t$, with $H$ the number of manually set hidden units. $f_t, i_t, o_t, c_t \in \mathbb{R}^{H \times 1}$ denote  the forget gate, input gate, output gate and updated cell state, respectively. $\sigma$  refers to the  sigmoid activation function \cite{APICELLA202114}.  $W_f, W_i, W_o, W_c \in \mathbb{R}^{H \times K}$, $U_f, U_i, U_o, U_c \in \mathbb{R}^{H \times H}$ and $V_f, V_i, V_o, V_c \in \mathbb{R}^{H \times K}$ represent  the weight matrices  of the current input $I_t$, previous output $h_{t-1}$ and  missing indicator $m_t$, respectively.  $b_f, b_i, b_o, b_c \in \mathbb{R}^{H \times 1}$ are the bias vectors for each function. All the weights and biases need to be learned. The LSTM-M network yields the hidden representation sequence $\{h_1,h_2,...,h_T \in \mathbb{R}^{H \times 1}\}$.

Three types  of attention modules -- inspired by how clinicians inspect EHR records -- will process the hidden representation sequence to extract different information.
\subsubsection{\textcolor{black}{General  attention}}

\textcolor{black}{When a physician accesses a patient's EHR, the initial step involves a comprehensive review of the entire record to gain a holistic understanding of the patient's medical history. This process is similar to how certain records may capture the physician's attention during their evaluation. Reflecting this approach, we  implement a general  attention mechanism to calculate  the interaction of individual
records. This mechanism is designed to assign an attention weight to each hidden representation in the following manner:}

% \begin{equation}
%     \left[\alpha_1,\alpha_2,...,\alpha_T\right]=softmax(\frac{h_t {h}^\mathrm{T}_1}{\sqrt{K}},\frac{h_t {h}^\mathrm{T}_2}{\sqrt{K}},...,\frac{h_t {h}^\mathrm{T}_T}{\sqrt{K}})  
% \end{equation}

\begin{equation}
    \left[\alpha_1,\alpha_2,...,\alpha_{T-1}\right]=softmax\left(\frac{ {h}^\top_T h_1}{\sqrt{K}},\frac{ {h}^\top_T h_2}{\sqrt{K}},...,\frac{ {h}^\top_Th_{T-1}}{\sqrt{K}}\right), 
\end{equation}

\begin{equation}
  h_{\alpha_T}=\sum_{t=1}^{T-1} \alpha_t h_{\mathrm{t}},
\end{equation}
where ${softmax}{(x)_i} = \frac{{{e^{{x_i}}}}}{{\sum\limits_{j = 1}^n {{e^{{x_j}}}} }}$.
The final hidden representation $h_{\alpha_T}$, which encompasses the information from all hidden representations, is obtained through the local attention module.
\subsubsection{\textcolor{black}{Feature-level attention}}
\textcolor{black}{Physicians often prioritize recent results when evaluating tests pertinent to the specific disease in question. It is important to recognize that the 'recent' of data may vary for each test, the phenomenon we described before as 'local irregularity'. To mimic this clinical approach, we develop a feature-level attention module. This module is specifically designed to consider the irregularity of each test item and efficiently extract significant insights from the hidden representations. First, all hidden representations are converted into a query vector, as follows:}
\begin{equation}
    h_{\star}=ReLU(W_h \left[ h_1,h_2,...,h_T \right]+b_h),
\end{equation}
where $W_h \in \mathbb{R}^{1 \times H}$ and $b_h \in \mathbb{R}^{1 \times T} $  are the parameters to be learned. $ReLU$ refers to the rectified linear unit activation functions \cite{APICELLA202114}. Next, we use the irregular time interval $\widetilde{e}_t$ as the key vector to calculate the attention in $h^{\star} \in \mathbb{R}^{1 \times T}$ and obtain the new hidden representation, given by:

% \begin{equation}
%     \left[\gamma_1, \gamma_2,...,\gamma_T\right]=softmax\left(\frac{\widetilde{e}_1 W_{\gamma} h_{\star}^\top}{\sqrt{K}},\frac{\widetilde{e}_2 W_{\gamma} h_{\star}^\top}{\sqrt{K}},...,\frac{\widetilde{e}_{T-1} W_{\gamma} h_{\star}^\top}{\sqrt{K}}\right)         
% \end{equation}

\begin{equation}
\begin{gathered}
\alpha_i = \frac{\widetilde{e}_i W_{\gamma} h_{\star}^\top}{\sqrt{K}}, \quad i = 1, 2, ..., T-1 \\
\left[\gamma_1, \gamma_2,...,\gamma_T\right] = softmax\left(\alpha_1, \alpha_2, ..., \alpha_{T-1}\right)
\end{gathered}
\end{equation}

\begin{equation}
  h_{\gamma_T}= \sum_{t=1}^{T-1}\gamma_t h_{\top},
\end{equation}
where $W_{\gamma} \in \mathbb{R}^{K \times T}$ is a parameter matrix to be learned. $h_{\gamma_T}$ represents  the hidden representation containing  information on local irregularity.
\subsubsection{\textcolor{black}{Label-level attention}}

\textcolor{black}{Beyond considering the irregular time intervals of individual test items (the feature-level attention), physicians also evaluate each  visit comprehensively. In this holistic assessment, earlier visits are typically accorded less attention (referred as the global irregularity).} To achieve this, we use label-level attention.  The time interval  $\delta_t$ is converted into a decay coefficient and combined with each hidden representation to obtain the global time-aware hidden representation, expressed as:
\begin{equation}
    \widetilde{\delta}_t=W_{\delta2}\left(1-tanh \left(\left(W_{\delta1} \frac{\delta_t}{\delta_{max}}+b_{\delta1}\right)^2\right)\right)+b_{\delta2},
\end{equation}
\begin{equation}
    \left[\beta_1,\beta_2,...,\beta_{T-1}\right]=softmax(\widetilde{\delta}_1,\widetilde{\delta}_2,...,\widetilde{\delta}_{t-1}), 
\end{equation}
\begin{equation}
  h_{\beta_T}= \sum_{t=1}^{T-1}\beta_t h_{\mathrm{t}},
\end{equation}
where $W_{\delta1}, W_{\delta2}, b_{\delta1}, b_{\delta2} \in \mathbb{R}$ are the parameters to be learned. 

Thus far, we have obtained the final  hidden representation $h_{\alpha T}$, $h_{\gamma T}$ and $h_{\beta T}$  from the local attention, feature-level attention and  label-level attention. They are finally  combined as $h_{\tau_ T}=h_{\alpha_T}+h_{\gamma_T}+h_{\beta_T}$ to form the input for downstream task-oriented modules.
\subsubsection{Down-stream task-oriented modules}
In this study, the downstream tasks involve predicting hemoglobin levels (regression) and anemia degree (classification). Using $h_{\tau_ T}$, a 3-layer Multilayer Perceptron (MLP) is used to perform these tasks. 
% The  tanh layer and the softmax layer are set as the final layer of the MLP for  each task, 
 Root mean squared error (RMSE) and multi-class cross-entropy  are selected  as  the respective loss functions for learning the parameters of the various modules described previously.
% \begin{itemize}
%     \item  Regression:
%     \begin{equation}
%     \sqrt{\frac{1}{N} \sum_{n=1}^N\left(\hat{y}^{(n)}-y^{(n)}\right)^2}
%     \end{equation}
%     \item Classification:
%     \begin{equation}
%     \frac{1}{N} \sum_{n=1}^N\left(\frac{1}{K} \sum_{k=1}^K-\beta\left(1-\hat{y}_k^{(n)}\right)^\gamma \log \left(\hat{y}_k^{(n)}\right)\right)
% \end{equation}
% \end{itemize}
% where $N$ is the number of the sample, $C$ is the number of all the labels, $\gamma$ and $\beta$ are the adjustable parameters, $\hat{y}^{(n)}$ is the ground truth label and $y^{(n)}$ is the predicted result.

% \subsubsection{Re-purposing the model from use case 1 to use case 2} The aforementioned  model description applies to our use case 1. Minor modifications are required for use case 2: add time point $T+1$ to the data input and duplicate  $x_T$, renaming  it as $x_{T+1}$. Replace the corresponding value in $x_{T+1}$ with the test result at $T+1$. All other values, such as $e_{T+1}$, $m_{T+1}$, and $\delta_{T+1}$, can be calculated, and the entire  process can be executed.

%% file: SPLITEE/chapters/Algorithm.tex
\section{Experimental setting}
\subsection{Dataset}
Two real-world datasets are chosen for our experiments: MIMIC III and eICU.

1) MIMIC III dataset

$\bullet$ Overview: The MIMIC III dataset comprises information on approximately 60,000 patients who stayed in critical care units at Beth Israel Deaconess Medical Center between 2001 and 2012 \cite{johnson2016mimic}. 
    % The dataset contains 26 tables, i.e. chartevents (all charted observations for patients) and diagnoses$\_$ICD (Hospital assigned diagnoses, coded using the International Statistical Classification of Diseases and Related Health Problems (ICD) system).
    
$\bullet$ Data selection: The selected data encompasses  five parts: 1) demographic information (18 kinds); 2) vital signs (652 kinds); 3) lab test results (753 kinds); 4) medication records (4525 kinds); 5)  ICD-9 code (14568 kinds). The hemoglobin value is extracted from lab test results and acts directly as the label for our regression problem (hemoglobin level prediction) and is used to calculate the label for our classification problem (anemia severity prediction). In particular, the anemia degree is calculated from the hemoglobin level using four different categories: 1) non-anemia 2) mild anemia 3) moderate anemia and 4) severe anemia, according to the WHO guidelines shown in Table \ref{Hemoglobin levels for diagnosing anemia} \cite{world2011haemoglobin}.  The details of selected features are shown in the Appendix.

%增加一下绘制直方图和表
    \begin{table}[htbp]
 	\centering
    \small
 	\setlength{\tabcolsep}{1mm}
 	\caption{Hemoglobin levels for diagnosing anemia (g/dL)}\label{Hemoglobin levels for diagnosing anemia}
 	\begin{tabular}{ccccc}
 		\hline
 		\tabincell{l}{Population} 
 		&\tabincell{c}{Non\\anemia} 
 		&\tabincell{c}{Mild \\anemia}
            &\tabincell{c}{Moderate \\anemia}
            &\tabincell{c}{Severe\\ anemia}
   
   \\ \hline
 		
 \tabincell{c}{age: 6-59 months} & $\geq$11.0 &10.0-10.9 & 7.0-9.9 &  $<$7.0\\
  \tabincell{c}{age: 5-11 years}& $\geq$11.5 &11.0-11.4 & 8.0-10.9 &  $<$8.0\\ \tabincell{c}{age: 12-14 years} & $\geq$12.0 &11.0-11.9 & 8.0-10.9 &  $<$8.0\\
 \tabincell{c}{Non-pregnant women \\(age: beyond 15 years)} & $\geq$12.0 &11.0-11.9 & 8.0-10.9 &  $<$8.0\\
 \tabincell{c}{Pregnant women} & $\geq$11.0 &10.0-10.9 & 7.0-9.9 &  $<$7.0\\
 \tabincell{c}{Men (age: beyond 15 years)} & $\geq$13.0 &11.0-12.9 & 8.0-10.9 &  $<$8.0\\
 \hline
 	\end{tabular}
 \end{table}
    
    $\bullet$ Data pre-processing: Test results with 100$\%$ missing values are removed.  One-hot encoding is applied to categorical  data and all data are normalized to have a variance of 1 and a mean of 0. Subsequently, the minimum number of records per patient is set to 80 to ensure adequate  information. As records from a long time ago provide  limited value  to current estimations, only records with intervals shorter than 7 days are kept.  Ultimately, a dataset containing  9620 samples is obtained.  %原本是60000(筛选出7504个,599968条数据和449652条训练样本)个病人，我们只用了8000个病人，得到了1025个病人的81952条病历，按照滑动窗口为20的话，我们得到了61452（20*35）条训练样本 ,
    %直接预测的话是使用了8000条数据来查看数据对比
    
    %在后面记得说一下baseline里面使用的插补方法

2) eICU dataset

    $\bullet$ Overview: The eICU dataset comprises  information from  139367 patients across  208 hospitals in America between 2014 and 2015 \cite{pollard2018eicu}. 
    
    $\bullet$ Data selection: The selected data contain five parts: 1) demographic (11 kinds);  2) vital signs (128 kinds); 3) lab test results (158 kinds);  4) medication records (1412 kinds); 5) ICD-9 code (1208 kinds). We determine the relevant label for our regression and classification problems by adopting the previous procedure. The details of selected features are shown in the Appendix. 
    
    $\bullet$ Data pre-processing: The pre-processing procedure is identical to that of the MIMIC III dataset. Ultimately, the dataset containing  11742 samples is obtained.  %在后面记得说一下baseline里面使用的插补方法

% The histograms of the hemoglobin level label of two datasets are shown in Fig. \ref{The histograms of the hemoglobin level label of MIMIC III dataset} and Fig. \ref{The histograms of the hemoglobin level label of eICU dataset}, respectively. 
The statistics of  anemia degree labels are shown in Table \ref{The statistics of the anemia degree label of two datasets}. The statistics of the preprocessed features and the missing rate are shown in  Table \ref{statistical information on the missing data}.

% \begin{figure}[htbp]
% 	\centering
% 	\includegraphics[width=3.5in]{Figure/The histograms of the hemoglobin level label of MIMIC III dataset.jpg}
% 	\caption{The histograms of the hemoglobin level label of MIMIC III dataset}
% 	\label{The histograms of the hemoglobin level label of MIMIC III dataset}
% \end{figure}
% \begin{figure}[htbp]
% 	\centering
% 	\includegraphics[width=3.5in]{Figure/The histograms of the hemoglobin level label of eICU dataset.jpg}
% 	\caption{The histograms of the hemoglobin level label of eICU dataset}
% 	\label{The histograms of the hemoglobin level label of eICU dataset}
% \end{figure}

\begin{table}[htbp]
 	\centering
        \small
 	\setlength{\tabcolsep}{2mm}
 	\caption{The statistics of the anemia degree label of two datasets}\label{The statistics of the anemia degree label of two datasets}
 	\begin{tabular}{ccc}
 		\hline
 		\tabincell{l}{Anemia degree label} 
 		&\tabincell{c}{MIMIC III dataset} 
 		&\tabincell{c}{eICU dataset}
    
   \\ \hline
 		
 \tabincell{c}{Non-anemia} & 5000 &2897 \\
  \tabincell{c}{Mild anemia}& 1858 &2781 \\ 
 \tabincell{c}{Moderate anemia} &2440  &5699 \\
 \tabincell{c}{Severe anemia} & 322 &365\\
 \tabincell{c}{Total} & 9620 &11742 \\
 \hline
 	\end{tabular}
 \end{table}

\begin{table}[h]
 	\centering
        \small
 	\setlength{\tabcolsep}{1mm}
 	\caption{\textcolor{black}{The statistical information on the missing data in the MIMIC III and eICU dataset. }}\label{statistical information on the missing data}
 	\begin{tabular}{ccccc}
 		\hline
   
 		&\multicolumn{2}{c}{MIMIC III} &\multicolumn{2}{c}{eICU}
   \\ 
   Feature category
   &\tabincell{c}{Feature\\number} & \tabincell{c}{Mean  missing\\ rate ($\%$)}   &\tabincell{c}{Feature\\number} & \tabincell{c}{Mean  missing \\ rate ($\%$)}\\ \hline
Demographics &   17 & 17.84&11&20.61 \\
Vital signs & 153 & 25.48&93&27.49 \\
Lab tests & 228 & 51.95&116&50.96 \\
Medication & 43 & 94.50 &58&96.91\\
ICD & 59 & 92.59&55&93.64 \\
All &500 & 51.14 &333 & 58.46 \\
 \hline
 	\end{tabular}
 \end{table}

\subsection{Baseline approaches}
The following methods are chosen as baselines for comparison:

    $\bullet$ LSTM: LSTM is a  fundamental  model dealing with time-series data\cite{hochreiter1997long}. In this study, we use a unidirectional LSTM as one of our baselines.
    
    $\bullet$ Time-aware LSTM (T-LSTM): T-LSTM is proposed for  clinical subtypes classification. It introduces a time-decay function to adjust the input of long-term memory of  previous units, modulating the impact of records from different time intervals on the current diagnosis \cite{baytas2017patient}.
    
     $\bullet$ Dual-attention time-aware gated recurrent unit (DATA-GRU): DATA-GRU is designed for mortality risk
prediction. It not only incorporates  a time-decay structure but also  applies distinct attention to imputed and observed data using an unreliability-aware attention mechanism, demonstrating strong performance in predicting patient mortality risk \cite{tan2020data}.

     $\bullet$ Attention-based time-aware LSTM (ATTAIN): ATTAIN is created for septic shock prediction. It addresses irregular time intervals by adjusting the long-term memory weights of several previous LSTM cells \cite{zhang2019attain}. 
     
     $\bullet$ Multi-integration attention module (MIAM): MIAM is a versatile  framework for processing EHR data, validated for predicting in-hospital mortality, length of stay, and phenotyping.It learns directly from irregular clinical time series, employing a multi-attention module to process observed values, missing indicators, and time intervals simultaneously, outperforming state-of-the-art methods \cite{lee2022multi}. 
     
     $\bullet$ Hierarchical time-aware attention networks (HiTANet): HiTANet is introduced  for risk prediction. It combines local and global attention analyses through a dynamic attention fusion component, yielding competitive results \cite{luo2020hitanet}.

\textcolor{black}{Our approach -- which was Inspired by \cite{luo2020hitanet} -- uses a structure that incorporates attention mechanisms following the sequential model. In particular, as described earlier, we designed three attention mechanisms that address global and local irregularities for irregular time series data. We also incorporated missing indicators, as advocated in \cite{lee2022multi}, since this can lead to promising results. In summary, HgbNet uses a number of additional mechanisms in relation to the state of the art in order to confront the various challenges posed by EHR data processing.}
\subsection{Performance Metrics}
For predicting hemoglobin levels, we use RMSE, mean absolute error (MAE), and R2 score as evaluation metrics. These metrics are calculated as follows:
\begin{equation}
   \text{RMSE} = \sqrt{\frac{\sum_{i=1}^{n} (y_i - \hat{y}_i)^2}{n}},
\end{equation}
\begin{equation}
\text{MAE} = \frac{\sum_{i=1}^{n} |y_i - \hat{y}_i|}{n},
\end{equation}
\begin{equation}
R^2 = 1 - \frac{\sum_{i=1}^{n} (y_i - \hat{y}_i)^2}{\sum_{i=1}^{n} (y_i - \bar{y})^2},
\end{equation}
where $y_i$ and $\hat{y}_i$ represent  the true value and  predicted value, respectively whereas $\bar{y}$ is the mean of the true values and $n$ is the number of instances.
Smaller RMSE and MAE and larger R2 values indicate more accurate predictions. Note that the R2 score ranges  from 0 to 1.

For predicting anemia degree, we select weighted precision, weighted recall, and weighted F1 score as evaluation metrics. These are given by
\textcolor{black}{
\begin{equation}
    \text{Weighted precision} =\frac{\sum_{k=1}^{K} n_k \times \text {Precision}_k}{\sum_{k=1}^{K} n_k},
\end{equation}
\begin{equation}
    \text{Weighted recall} =\frac{\sum_{k=1}^{K} n_k \times \text {Recall}_k}{\sum_{k=1}^{K} n_k},
\end{equation}
\begin{equation}
    \text{Weighted F1} =\frac{\sum_{k=1}^{K} n_k \times \text {F1}_k}{\sum_{k=1}^{K} n_k}, 
\end{equation}
where $\text{TP}_k$, $\text{FP}_k$ and $\text{FN}_k$ denote  the number of instances correctly predicted as class $k$, instances from other classes incorrectly predicted as class $k$, and instances from class $k$ incorrectly predicted as other classes, respectively, i.e. $\text{Precision}_k = \frac{\text{TP}_k}{\text{TP}_k + \text{FP}_k}$, $\text{Recall}_k = \frac{\text{TP}_k}{\text{TP}_k + \text{FN}_k}$, and $\text{F1}_k = 2 \times \frac{\text{Precision}_k \times \text{Recall}_k}{\text{Precision}_k + \text{Recall}_k}$.  $K$ is the total number of classes and $n_k$ is the number of instances belonging to class $k$.} Precision and recall are commonly used  in classification problems and F1 score takes both aspects into account. All three metrics range from 0 to 1, with larger values indicating better results. Additionally, the evaluation metric in each category is weighted  to calculate the final metric, accounting for the impact of sample size in different categories.
%下面这些是weighted的计算
% \text{Weighted Precision} = \frac{\sum_{k=1}^{K} n_k \times \text{Precision}_k}{\sum_{k=1}^{K} n_k}

% \text{Weighted Recall} = \frac{\sum_{k=1}^{K} n_k \times \text{Recall}_k}{\sum_{k=1}^{K} n_k}

% \text{Weighted F1 Score} = \frac{\sum_{k=1}^{K} n_k \times \text{F1 Score}_k}{\sum_{k=1}^{K} n_k}
\subsection{Additional experimental procedures}

To address the issue of uneven sample categories, we employ 5-group-fold cross-validation (ensuring a fixed percentage of data in each class for each fold) to evaluate all models \cite{scikit-learn}. The training, validation, and test data are split in a 0.72: 0.08: 0.2 ratio. For models that require data imputation, we follow the methods described in the original papers, such as mean or nearest neighbor imputation \cite{tan2020data,luo2020hitanet}.
We first establish a hyperparameter range for grid search and determine the optimal parameters as follows:  $H=128$, $T=80$,  the batch size is 512, and the time step (the first dimension of $x_t$) is 40. The Adam optimizer, with an initial learning rate of 0.001, is chosen  for both tasks.  
Each training session consists of 500 epochs with early stopping (patience value set at 10). 
The experiments are conducted on 2 Tesla A100 GPUs with an Intel Xeon Gold 6248 CPU, using the CentOS 8 operating system and Python 3.6 interpreter.

%% file: SPLITEE/chapters/Experiments.tex
\section{Results}\label{sec: experiments}

We now summarize our results associated with the two proposed use cases.
\subsection{Use case 1}
\textcolor{black}{In this scenario, we predict a patient's  hemoglobin level/anemia degree at moment $T+1$ by using the data from moment 1 to moment $T$ (note that $T+1$ here refers to the next moment, which may be a few hours/days later)}. 
\subsubsection{Overall Performance}
\begin{figure*}[htbp]
\centering
\subfigure[LSTM]{
\includegraphics[width=4.7cm]{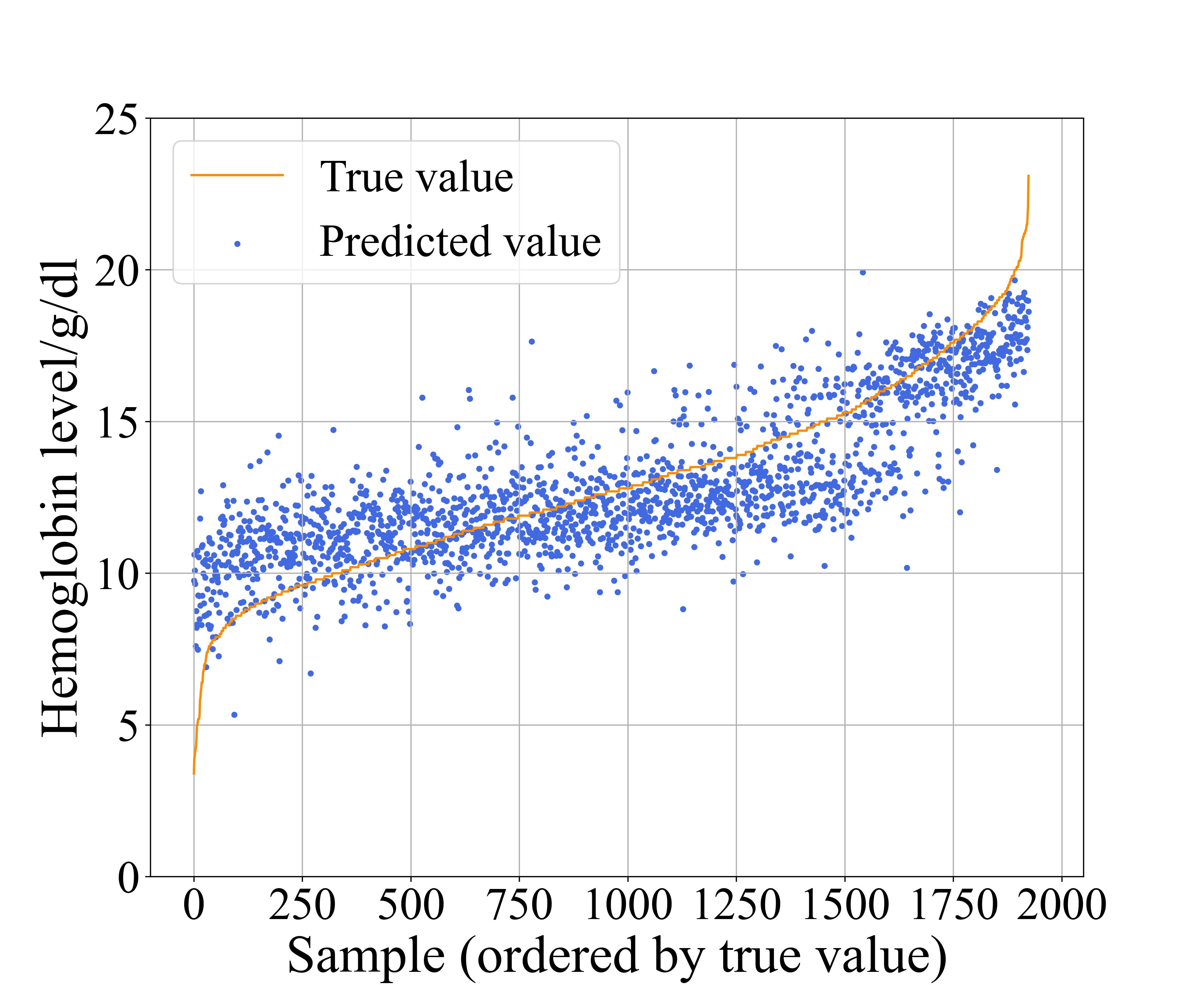}
%\caption{fig1}
}\hspace{-7mm}
\subfigure[T-LSTM]{
\includegraphics[width=4.7cm]{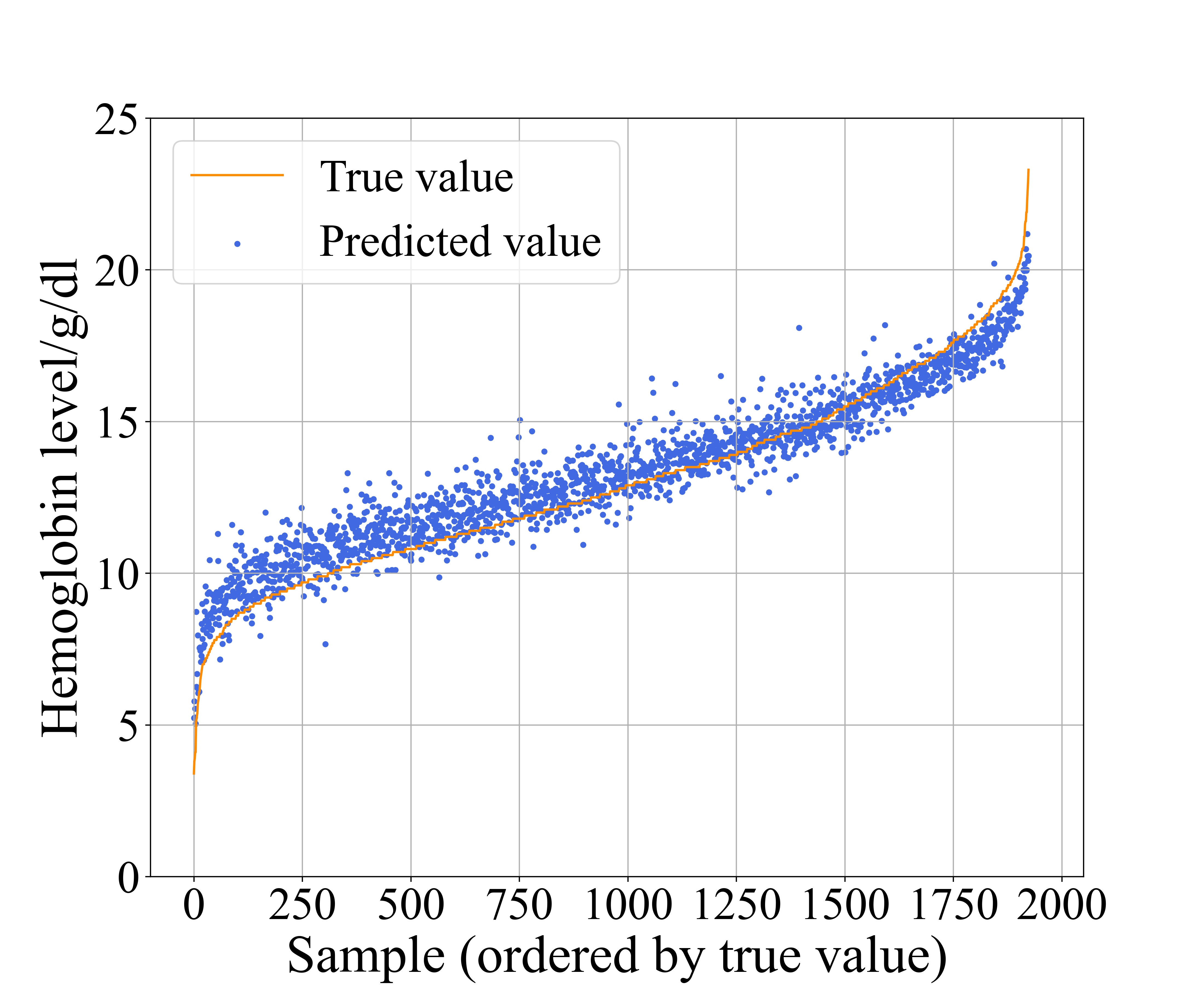}
%\caption{fig1}
}\hspace{-7mm}
\subfigure[Data-GRU]{
\includegraphics[width=4.7cm]{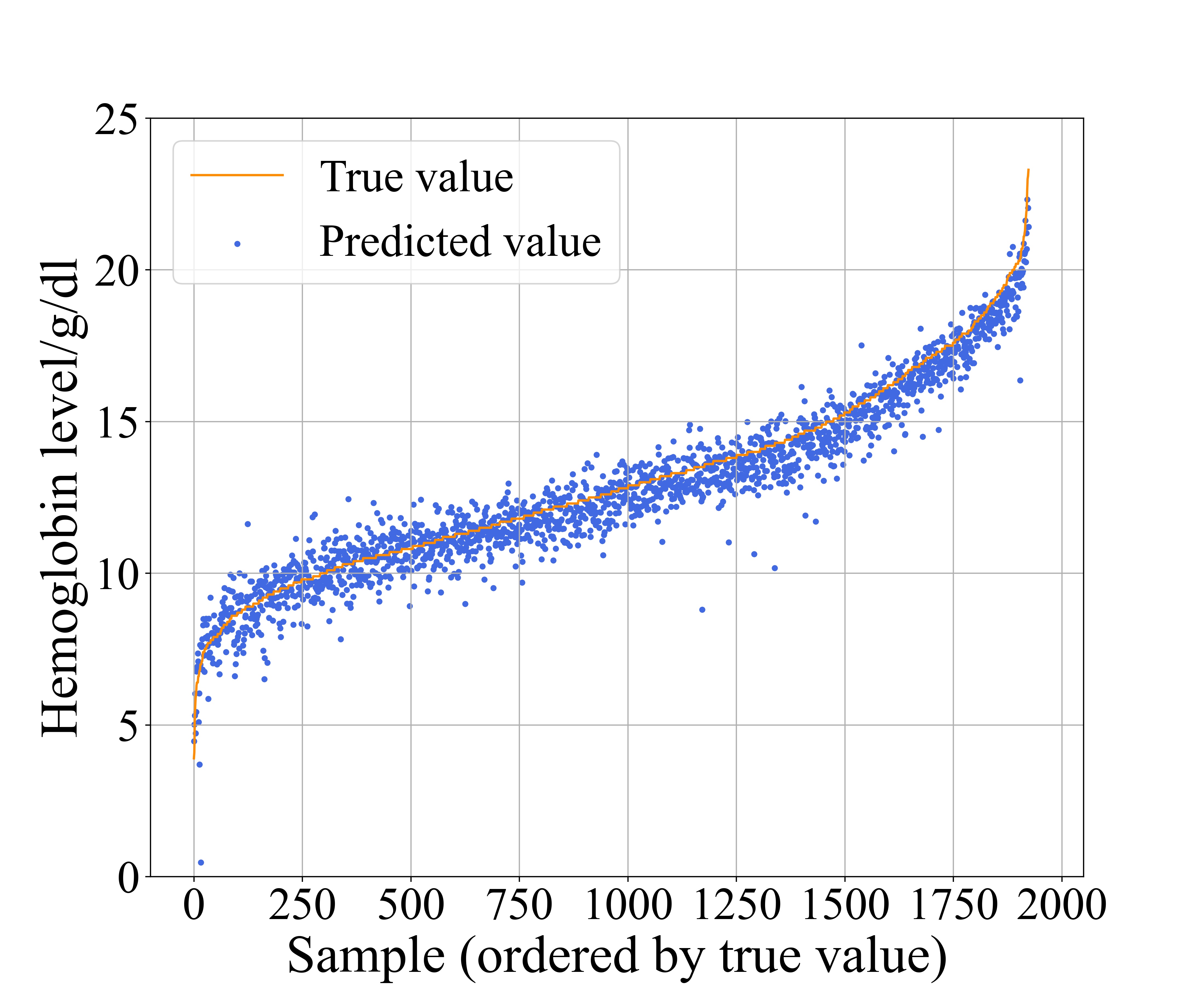}
%\caption{fig1}
}\hspace{-7mm}
\subfigure[ATTAIN]{
\includegraphics[width=4.7cm]{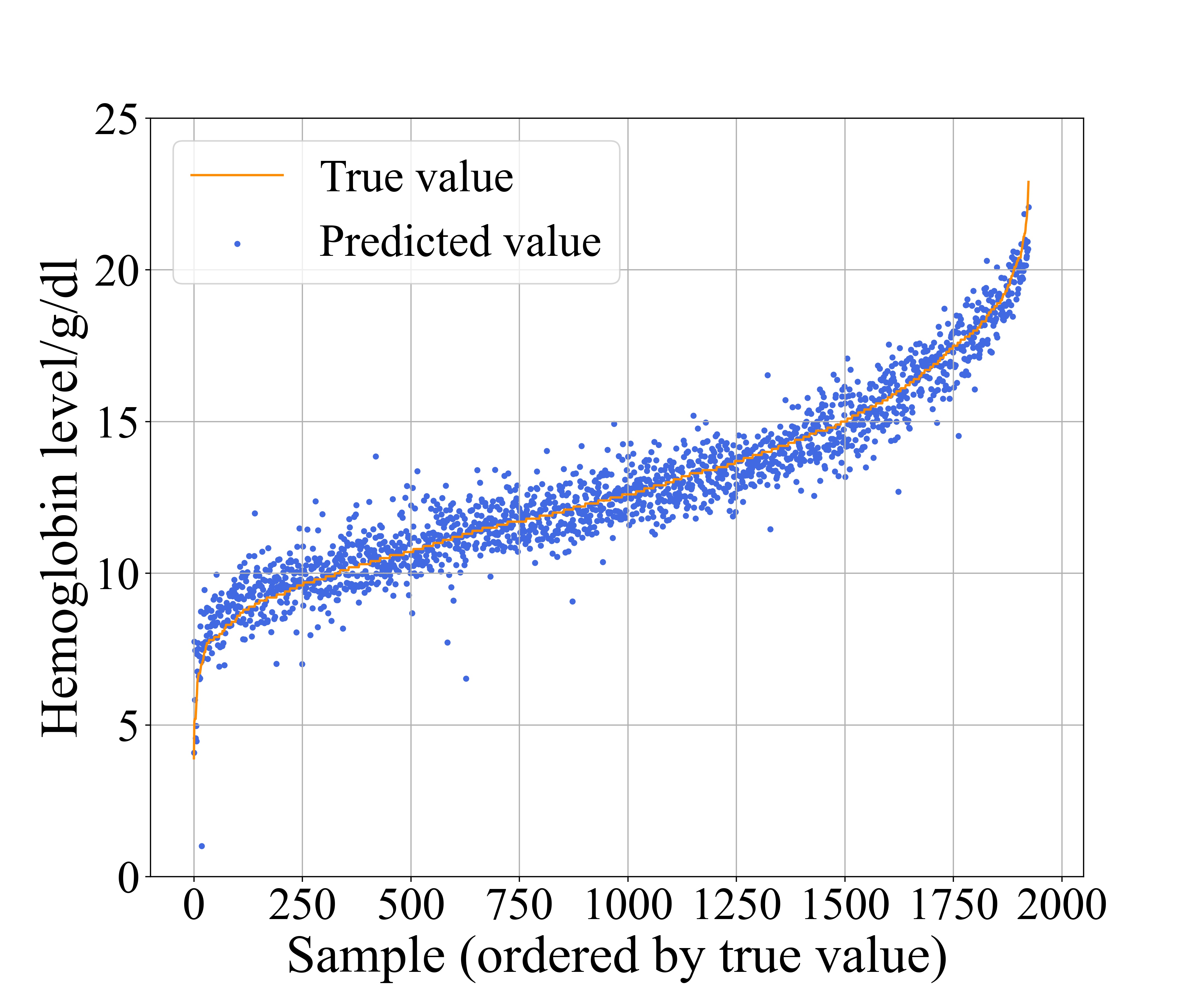}
%\caption{fig1}
}
\subfigure[MIAM]{
\includegraphics[width=4.7cm]{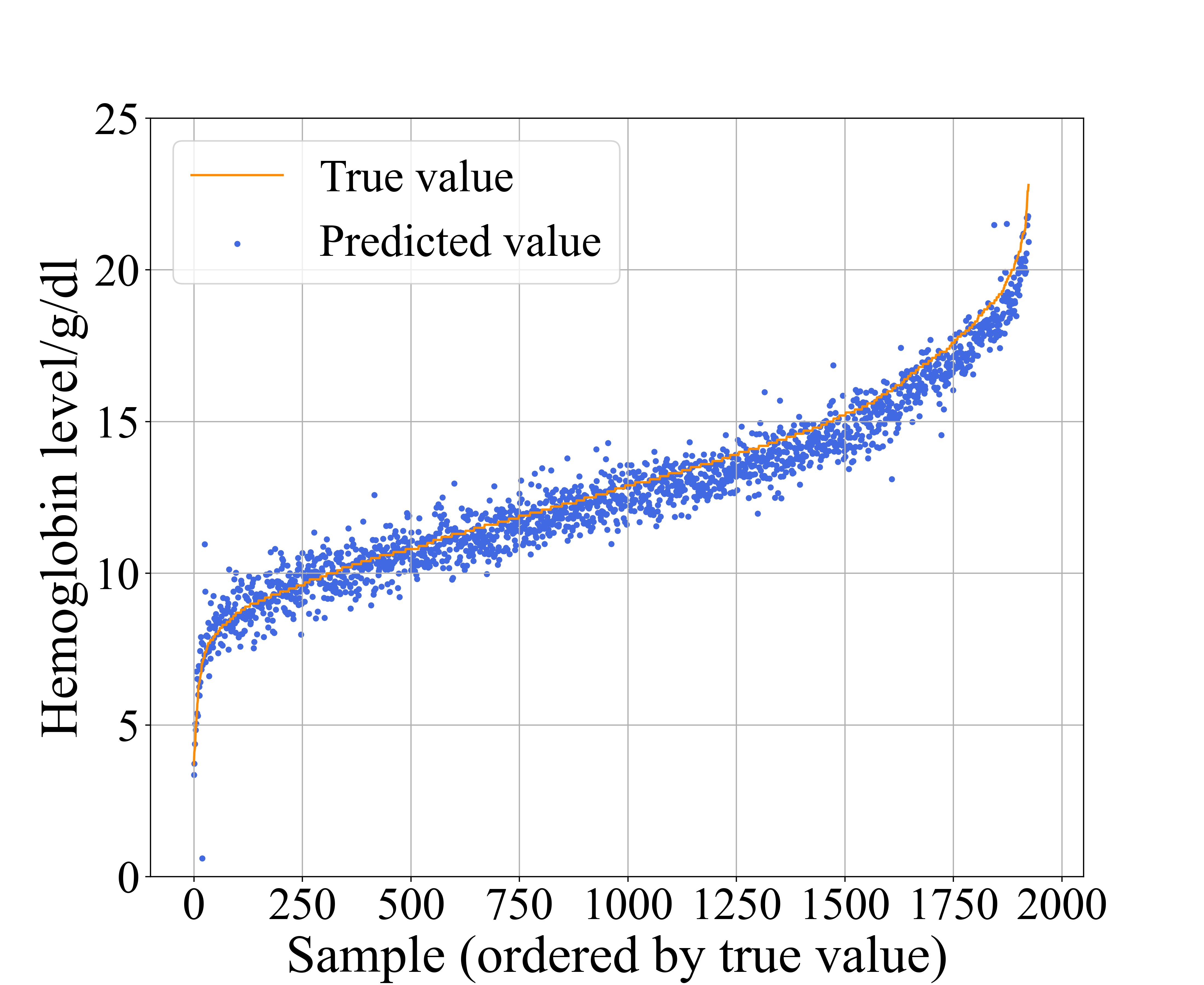}
%\caption{fig1}
}\hspace{-7mm}
\subfigure[HiTANet]{
\includegraphics[width=4.7cm]{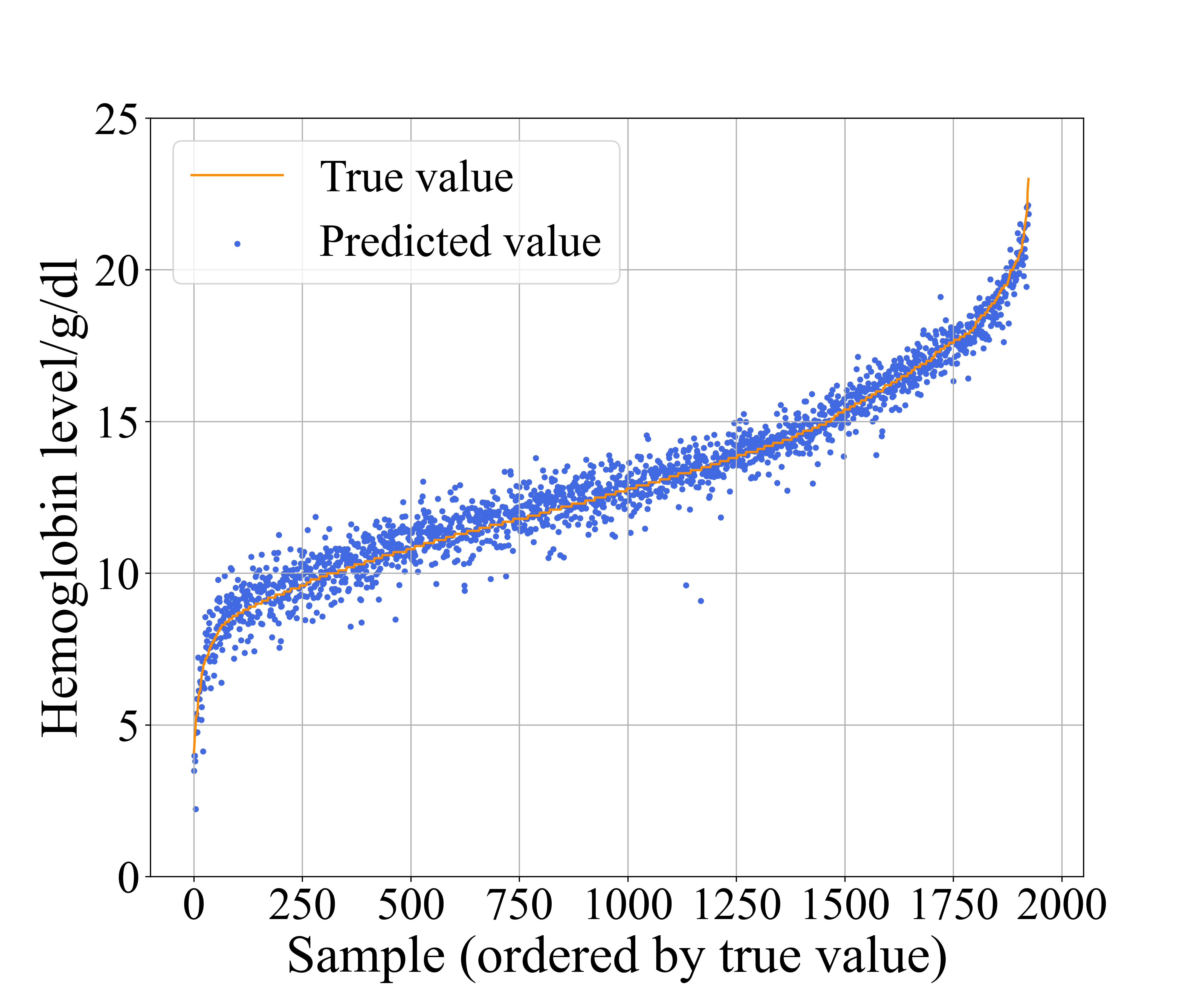}
%\caption{fig1}
}\hspace{-7mm}
\subfigure[HgbNet]{
\includegraphics[width=4.7cm]{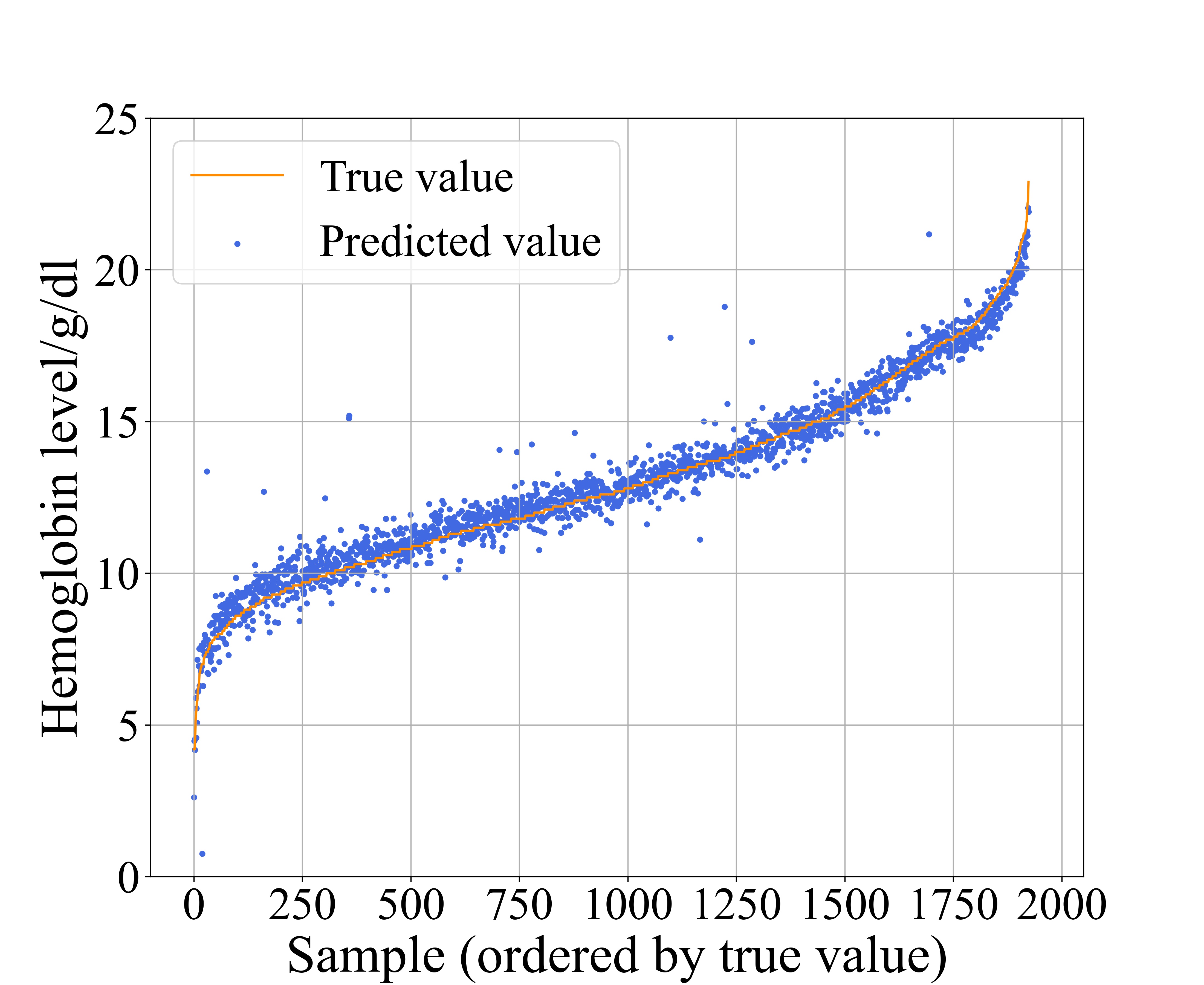}
%\caption{fig1}
}

\centering
\caption{The hemoglobin level prediction result of MIMIC III dataset}
\label{The hemoglobin level prediction result of MIMIC III dataset}
\end{figure*}

\begin{figure*}[htbp]
\centering
\subfigure[LSTM]{
\includegraphics[width=4.7cm]{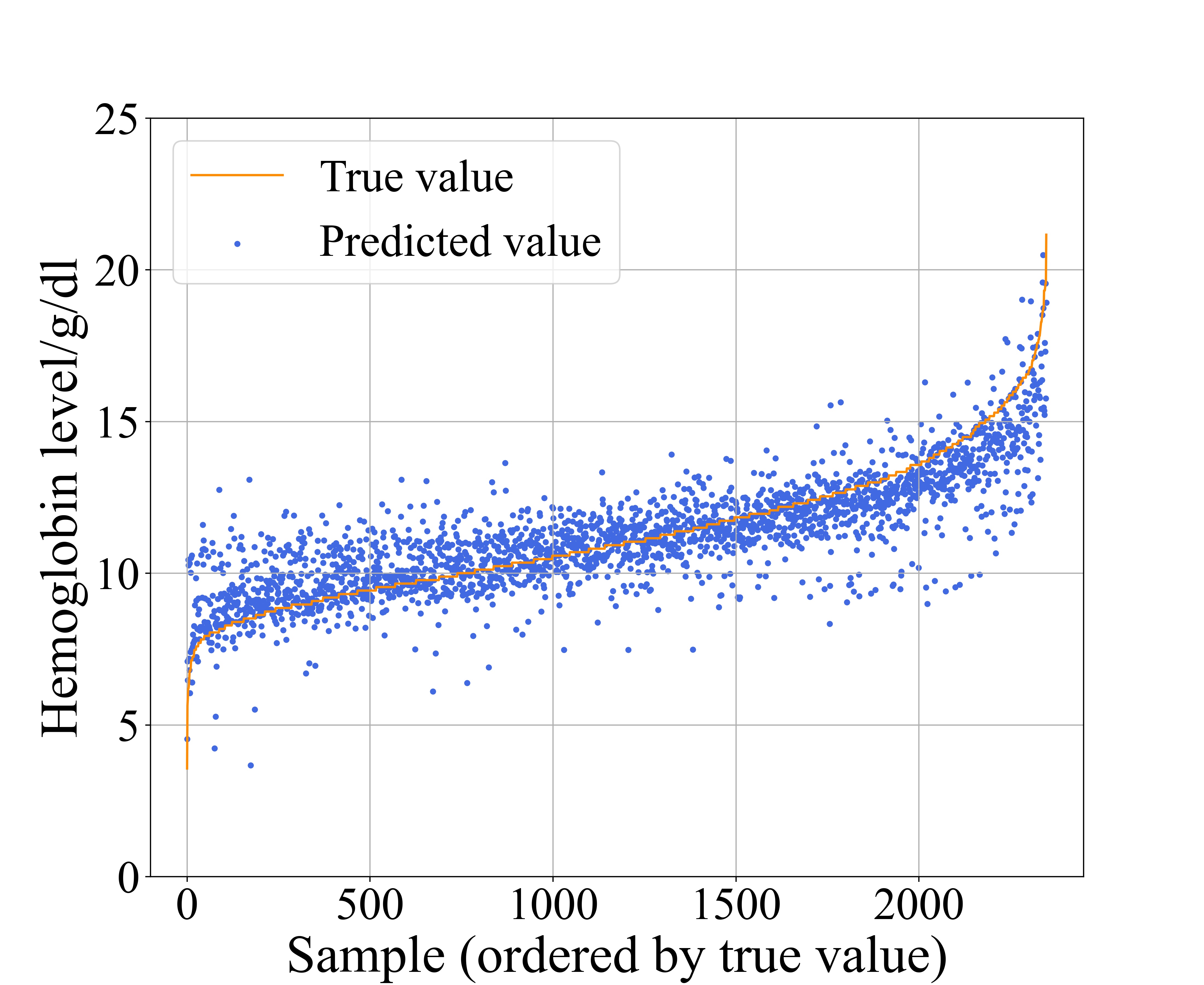}
%\caption{fig1}
}\hspace{-7mm}
\subfigure[T-LSTM]{
\includegraphics[width=4.7cm]{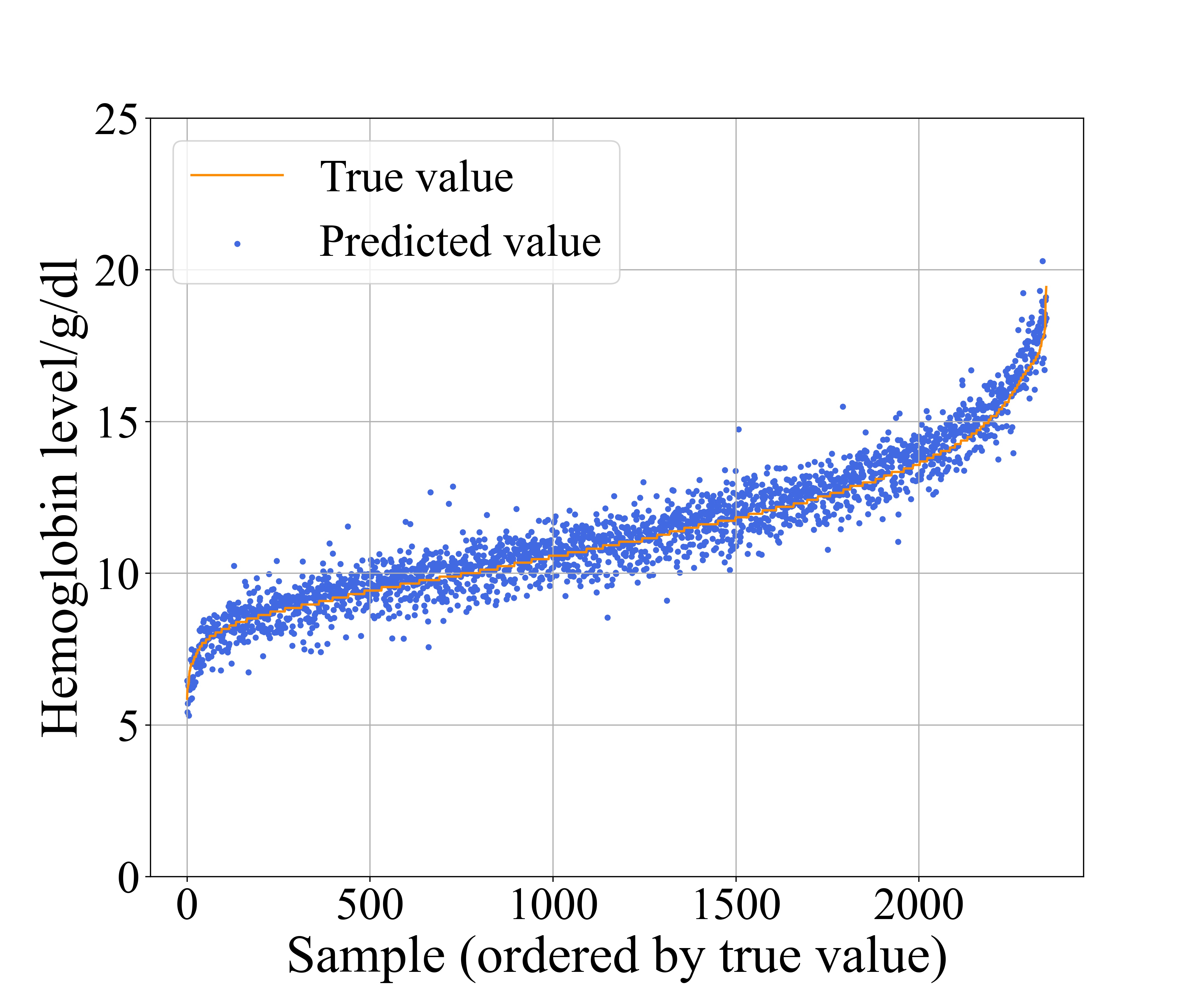}
%\caption{fig1}
}\hspace{-7mm}
\subfigure[Data-GRU]{
\includegraphics[width=4.7cm]{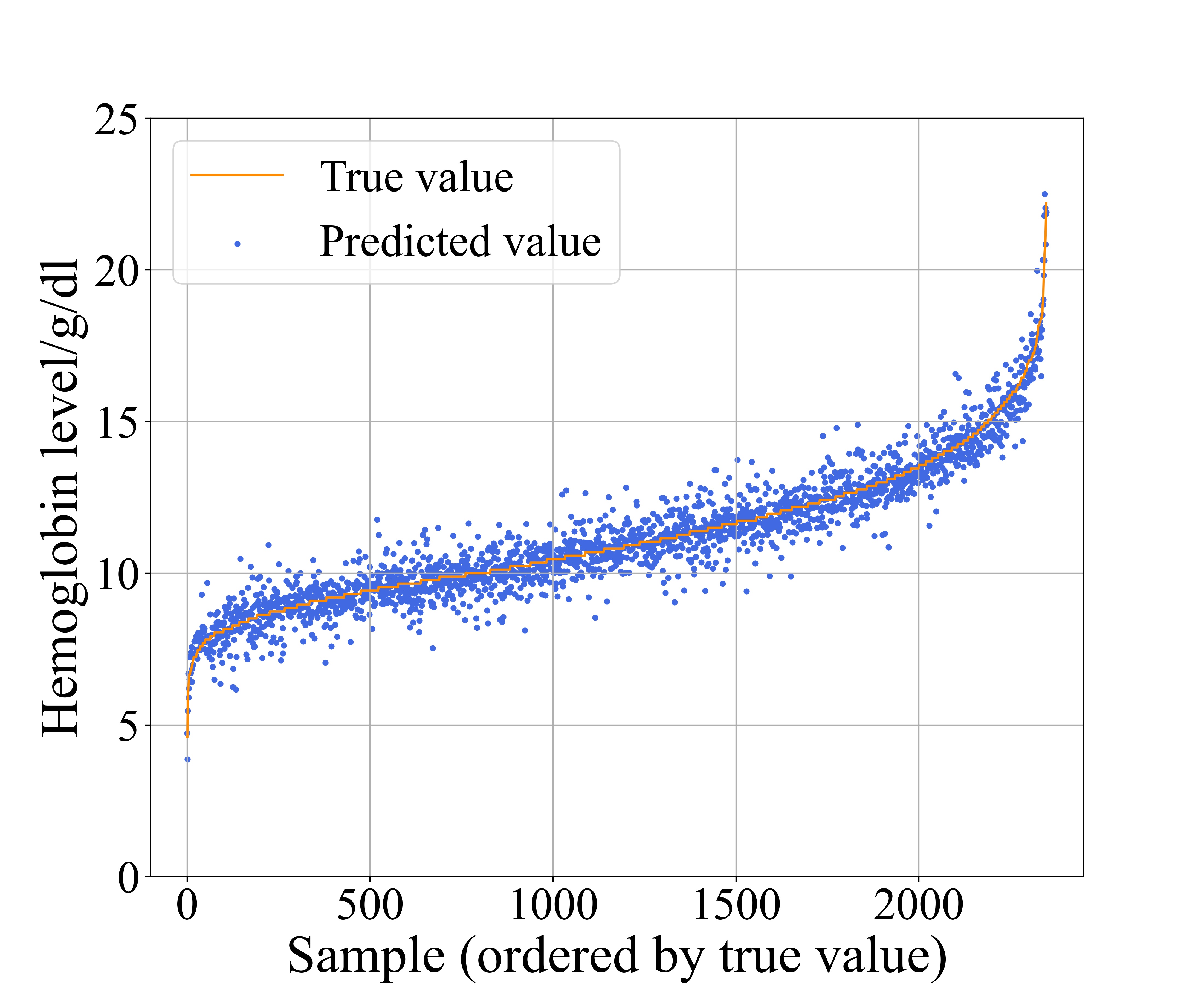}
%\caption{fig1}
}\hspace{-7mm}
\subfigure[ATTAIN]{
\includegraphics[width=4.7cm]{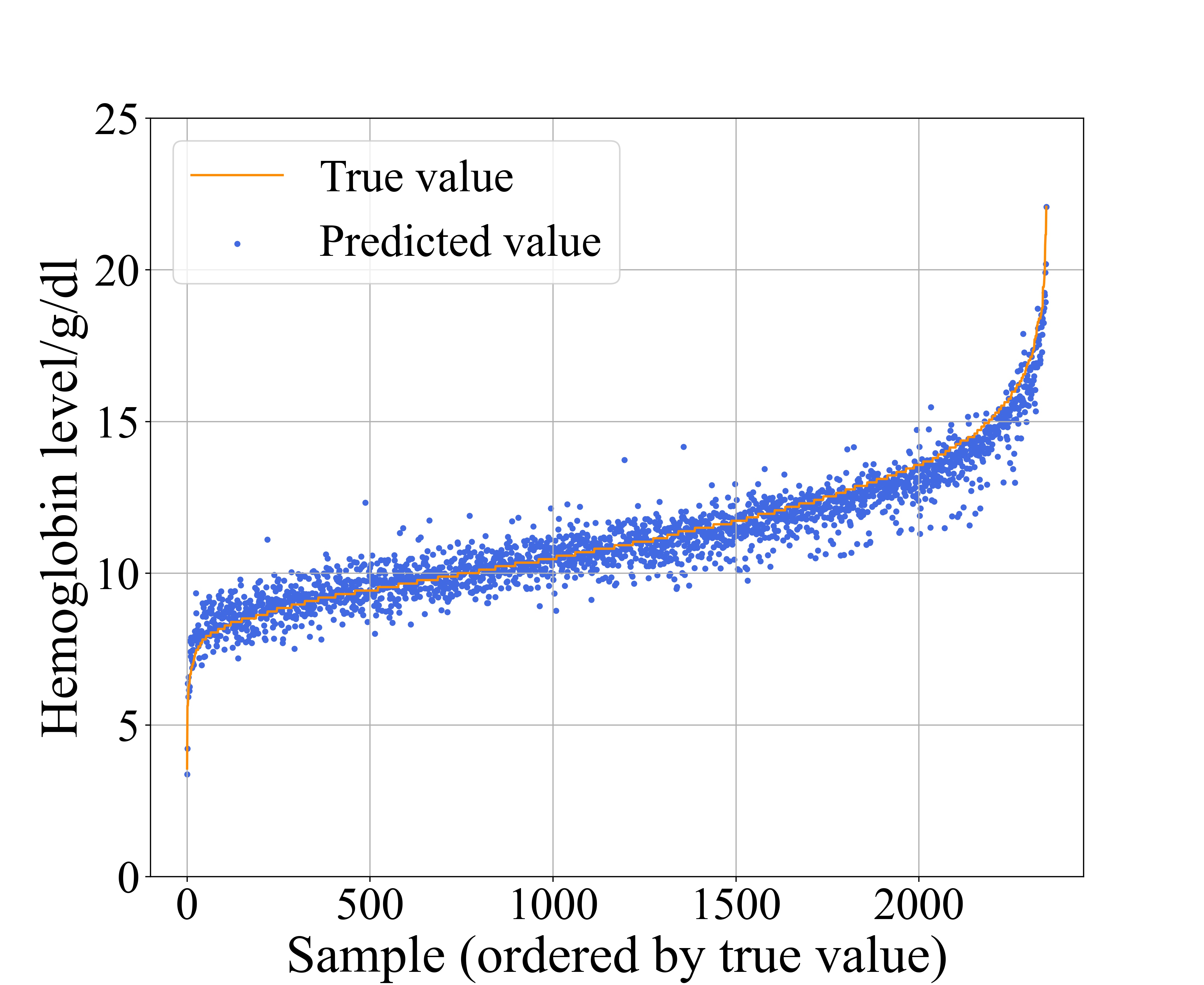}
%\caption{fig1}
}

\subfigure[MIAM]{
\includegraphics[width=4.7cm]{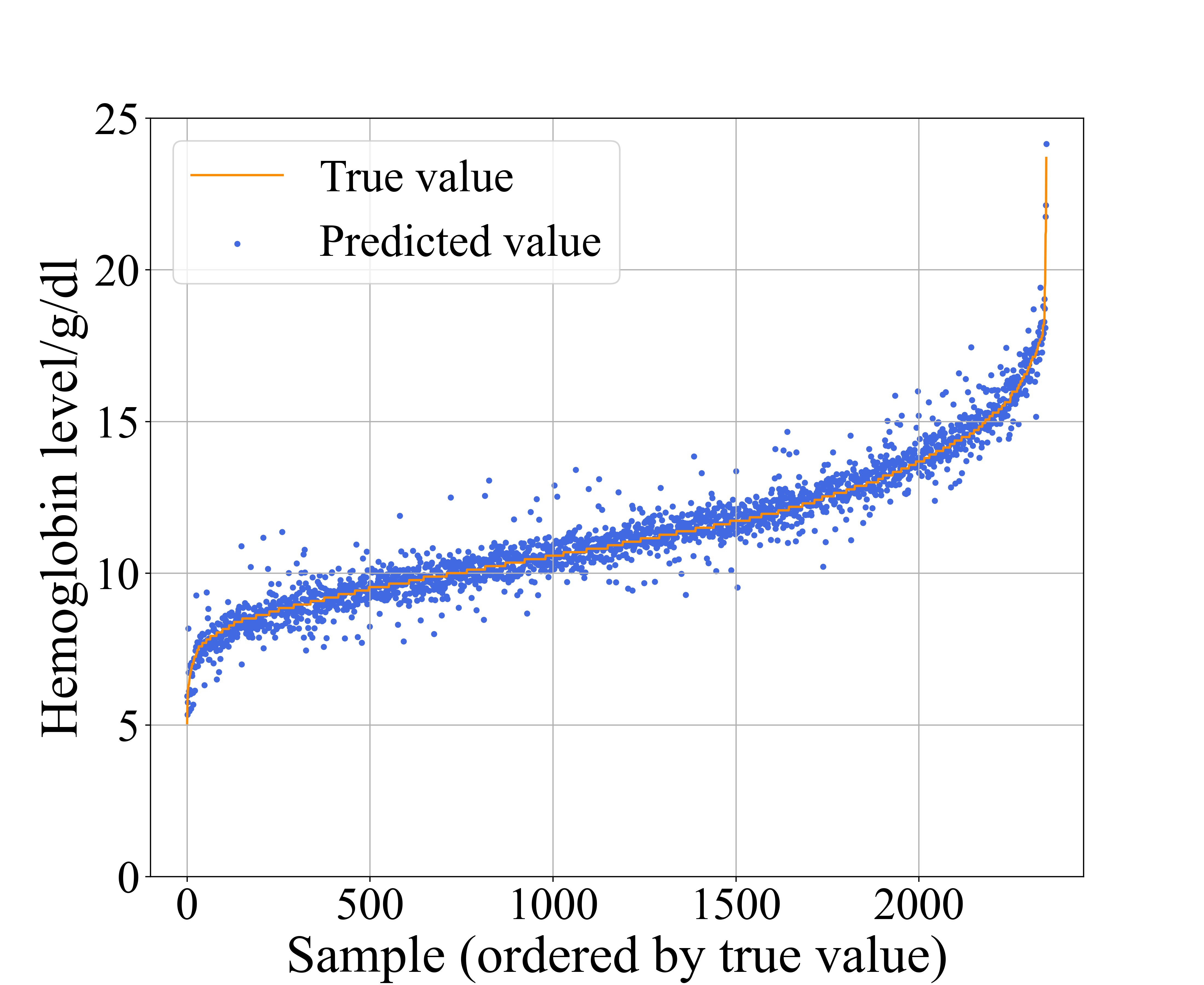}
%\caption{fig1}
}\hspace{-7mm}
\subfigure[HiTANet]{
\includegraphics[width=4.7cm]{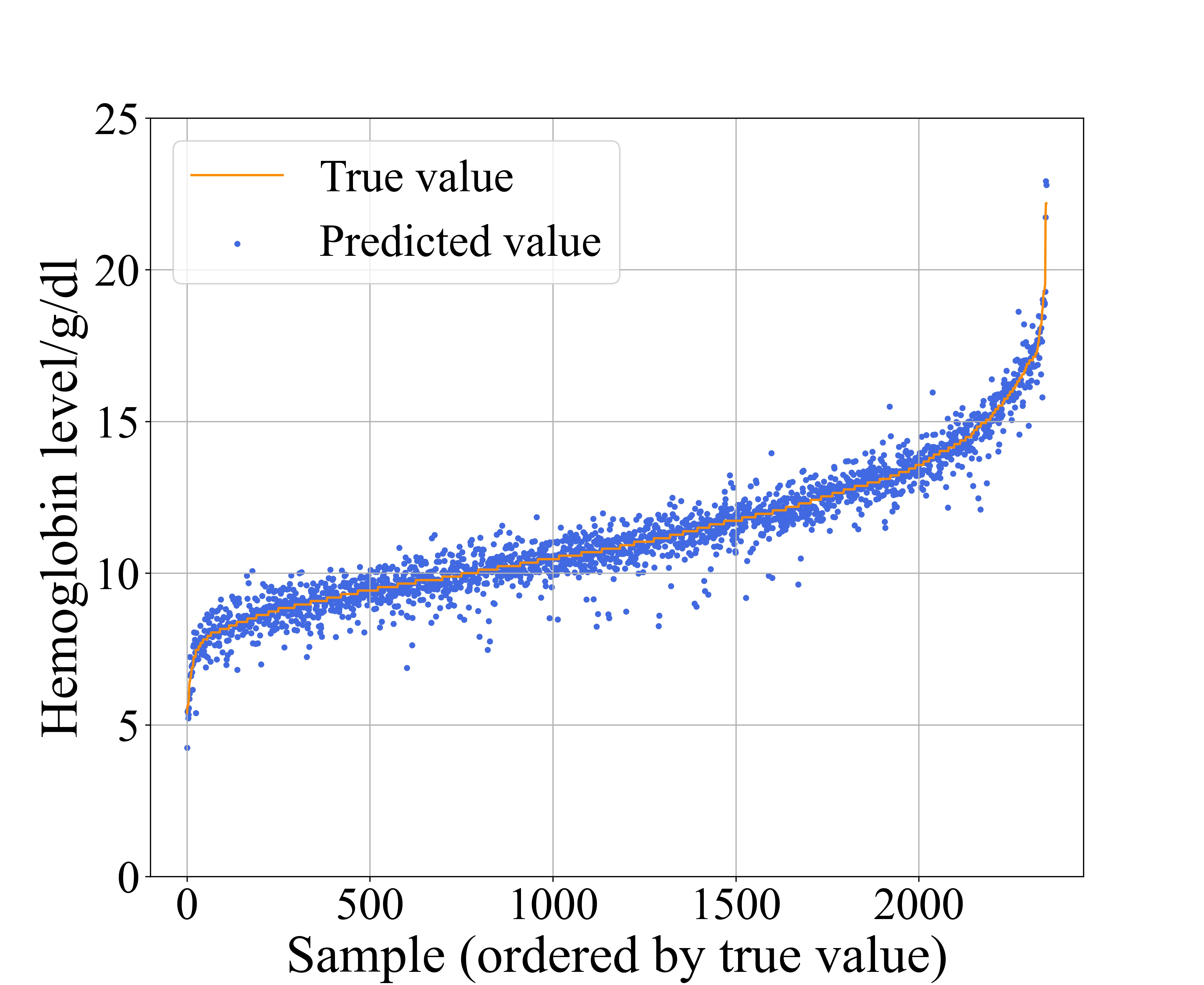}
%\caption{fig1}
}\hspace{-7mm}
\subfigure[HgbNet]{
\includegraphics[width=4.7cm]{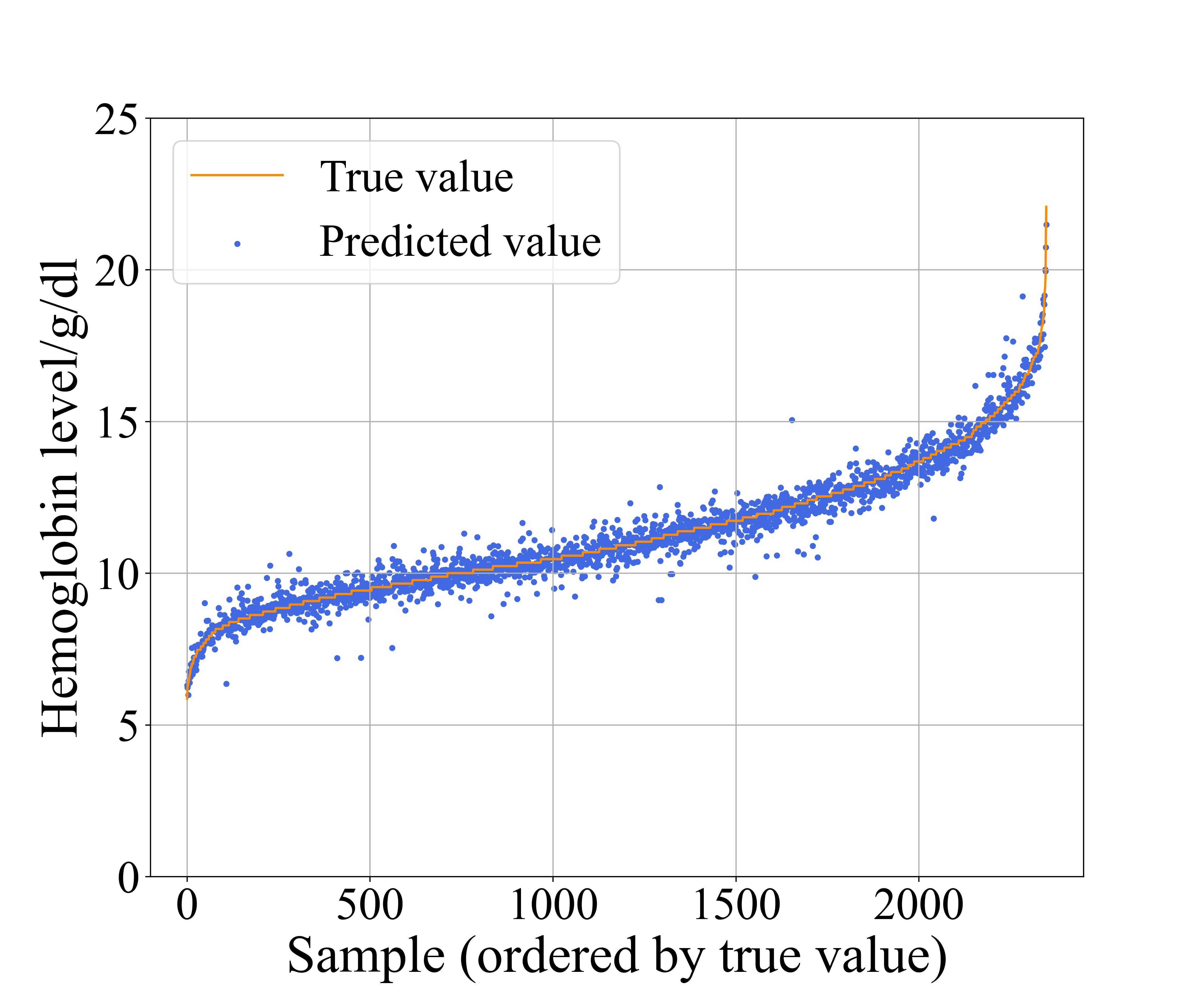}
%\caption{fig1}
}

\centering
\caption{The hemoglobin level prediction result of eICU dataset}
\label{The hemoglobin level prediction result of eICU dataset}
\end{figure*}

 Initially, we  plot the predicted hemoglobin levels against their corresponding true values, as illustrated in Fig. \ref{The hemoglobin level prediction result of MIMIC III dataset} and Fig. \ref{The hemoglobin level prediction result of eICU dataset} (samples are sorted by true value  for easy   observation). The evaluation matrix with corresponding standard deviation for hemoglobin level and anemia degree prediction are provided  in Table \ref{The evaluation matrix for hemoglobin level estimation.} and Table \ref{The evaluation matrix for anemia degree estimation.}.%每个模型的结果都是五折交叉验证的均值，那么我们画的这个是结果最接近均值的一次，前面要提一嘴就是为了保证样本的均匀，我们是用group-fold选取实验，这样可以保证结果更为均匀（每一个类分到的比例是固定的），

%mimic iii的hemoglobin排序图

%eICU的hemoglobin排序图

%考虑加上正负号，表示误差
%hemoglobin预测结果的统计表
\begin{table*}[htbp]\footnotesize
	\centering
	\setlength{\tabcolsep}{2mm}
	\caption{The evaluation matrix for hemoglobin level estimation.}\label{The evaluation matrix for hemoglobin level estimation.}
	\begin{tabular}{ccccccc}
		\hline
		Dataset  & \multicolumn{3}{c}{MIMIC III} & \multicolumn{3}{c}{eICU}\\ \hline
		\diagbox[dir=NW,height=2em,trim=r]{\footnotesize Model}{\footnotesize Metrics}             & RMSE     & MAE     & R2 score     & RMSE     & MAE     & R2 score     \\ \hline
LSTM       & 2.258 $\pm$\ 0.016   &1.791 $\pm$\ 0.014   & 0.685 $\pm$\ 0.010   & 1.663 $\pm$\ 0.014 & 1.287 $\pm$\ 0.012  &0.721 $\pm$\ 0.009 \\
T-LSTM     & 1.738 $\pm$\ 0.012   &1.339 $\pm$\ 0.011   & 0.805 $\pm$\ 0.007   &1.225 $\pm$\ 0.012  & 0.970 $\pm$\ 0.010  &0.810 $\pm$\ 0.005\\
Data-GRU   & 1.374 $\pm$\ 0.010  &1.030 $\pm$\ 0.010   & 0.843 $\pm$\ 0.004    &1.195 $\pm$\ 0.011 & 0.887 $\pm$\ 0.010  &0.822 $\pm$\ 0.003\\
ATTAIN     & 1.438 $\pm$\ 0.009   &1.097 $\pm$\ 0.010   & 0.836 $\pm$\ 0.004   &1.147 $\pm$\ 0.010 & 0.835 $\pm$\ 0.009  &0.828  $\pm$\ 0.004 \\
MIAM       & 1.310 $\pm$\ 0.008   &1.022 $\pm$\ 0.007   & 0.849 $\pm$\ 0.003  &1.005 $\pm$\ 0.010 & 0.690 $\pm$\ 0.008  &0.845  $\pm$\ 0.004\\ 
HiTANet    & 1.135 $\pm$\ 0.009   &0.850 $\pm$\ 0.009   & 0.862 $\pm$\ 0.004   &1.011 $\pm$\ 0.011 & 0.696 $\pm$\ 0.009  &0.843 $\pm$\ 0.003\\ 
HgbNet  & \textbf{1.076} $\pm$\ 0.010   &\textbf{0.712} $\pm$\ 0.008   & \textbf{0.867} $\pm$\ 0.003  &\textbf{0.809} $\pm$\ 0.008 & \textbf{0.646} $\pm$\ 0.007  &\textbf{0.861} $\pm$\ 0.003\\ 

%elcu的r2score会低一些但是rmse mae会好一些，因为杂点少,这里考虑要不要把这两个数据集的rmse和mae给换一下或者是把rmse和mae再调大一些
%这个RMSE和mae太小了，考虑乘1.5
\hline
	\end{tabular}
\end{table*}

%anemia的结果统计表
\begin{table*}[htbp]\footnotesize
	\centering
	\setlength{\tabcolsep}{2mm}
	\caption{The evaluation matrix for anemia degree estimation.}\label{The evaluation matrix for anemia degree estimation.}
	\begin{tabular}{ccccccc}
		\hline
		Dataset & \multicolumn{3}{c}{MIMIC III} & \multicolumn{3}{c}{eICU}\\ \hline
		\diagbox[dir=NW,height=2em,trim=r]{\footnotesize Model}{\footnotesize Metrics}

  & \tabincell{c}{Weighted\\Precision}  & \tabincell{c}{Weighted\\Recall}  &\tabincell{c}{Weighted\\F1 score}  & \tabincell{c}{Weighted\\Precision}  & \tabincell{c}{Weighted\\Recall}& \tabincell{c}{Weighted\\F1 score}   \\ \hline
  
LSTM   
& \tabincell{c}{0.646 $\pm$\ 0.013 }  & \tabincell{c}{0.651 $\pm$\ 0.011}& \tabincell{c}{0.639 $\pm$\ 0.010}
  &0.675  $\pm$\ 0.012  & 0.676  $\pm$\ 0.014  & 0.671 $\pm$\ 0.011\\ 

T-LSTM   & 0.798 $\pm$\ 0.009   &  0.822 $\pm$\ 0.008   &  0.799 $\pm$\ 0.008  &0.768 $\pm$\ 0.008  &   0.795 $\pm$\ 0.010 &    0.770  $\pm$\ 0.007 \\      

Data-GRU   

& \tabincell{c}{ 0.829 $\pm$\ 0.007} 
& \tabincell{c}{ 0.829 $\pm$\ 0.007}
& \tabincell{c}{0.829 $\pm$\ 0.006}
&0.830 $\pm$\ 0.007  &  0.840 $\pm$\ 0.008  &  0.824   $\pm$\ 0.005  \\

ATTAIN  & \tabincell{c}{0.828 $\pm$\ 0.006} 
& \tabincell{c}{0.838  $\pm$\ 0.008} 
& \tabincell{c}{0.817 $\pm$\ 0.006}   
&0.816  $\pm$\ 0.007  & 0.837 $\pm$\ 0.007  &  0.815    $\pm$\ 0.006 
\\

MIAM   
  & \tabincell{c}{0.836 $\pm$\ 0.007 }
& \tabincell{c}{0.837 $\pm$\ 0.006}
& \tabincell{c}{0.836 $\pm$\ 0.005}
&0.841  $\pm$\ 0.006 &  \textbf{0.853} $\pm$\ 0.006  &  0.833 $\pm$\ 0.007
 \\

HiTANet    & \tabincell{c}{0.819 $\pm$\ 0.006}
& \tabincell{c}{\textbf{0.885 }$\pm$\ 0.005}
& \tabincell{c}{0.843 $\pm$\ 0.006}
 &0.833 $\pm$\ 0.007  &   0.832 $\pm$\ 0.005 &   0.831 $\pm$\ 0.006  \\

HgbNet  &\tabincell{c}{ \textbf{0.859} $\pm$\ 0.006} 
&\tabincell{c}{ 0.854 $\pm$\ 0.006} 
&\tabincell{c}{ \textbf{0.855} $\pm$\ 0.005}
&\textbf{0.843} $\pm$\ 0.006 & 0.843 $\pm$\ 0.006 &  \textbf{0.843}  $\pm$\ 0.005  \\  

\hline
	\end{tabular}
\end{table*}

As illustrated in Fig. \ref{The hemoglobin level prediction result of MIMIC III dataset} and  Fig. \ref{The hemoglobin level prediction result of eICU dataset}, we can visually assess model performance in predicting hemoglobin levels. By reordering the true values according to their value and plotting the corresponding predicted values, we gain insights into three key performance aspects:

    $\bullet$ General prediction performance: The dispersion of the blue points visually demonstrates the overall prediction accuracy. For instance, LSTM performs suboptimally on both datasets, as evidenced by the blue points scarcely adhering to the orange line trend in Fig. \ref{The hemoglobin level prediction result of MIMIC III dataset} (a) and  Fig. \ref{The hemoglobin level prediction result of eICU dataset} (a).
    
    $\bullet$ Prediction bias across intervals: For example, the T-LSTM's prediction in Fig. \ref{The hemoglobin level prediction result of MIMIC III dataset} (b)  \textcolor{black}{largely follows the true values' trend but significantly overestimates the low and mid hemoglobin level between 0 and 1500 sampling points, while underestimating the high hemoglobin level for the remaining points}.   This behavior may be attributed to the model's limited capacity to handle uneven sample distribution. 
    
    $\bullet$ Noise in predicted values: Notably, several models yield more noise in the MIMIC III dataset than the eICU dataset, as seen in Fig. \ref{The hemoglobin level prediction result of MIMIC III dataset} (g) versus   Fig. \ref{The hemoglobin level prediction result of eICU dataset} (g), and Fig. \ref{The hemoglobin level prediction result of MIMIC III dataset} (d) compared to  Fig. \ref{The hemoglobin level prediction result of eICU dataset} (d). Such noise may affect noise-sensitive evaluation metrics like RMSE, potentially due to dataset noise. 

Table \ref{The evaluation matrix for hemoglobin level estimation.} offers the  quantitative  assessment of hemoglobin level predictions. LSTM is the least effective, with the highest RMSE, MAE, and lowest R2 scores on both datasets.  This deficiency can be attributed to the LSTM model's inability to manage irregular time intervals and missing values. T-LSTM demonstrates improvement but underperforms compared to other baselines.  This outcome can be explained by T-LSTM's focus on irregular time intervals between visits while disregarding missing values.  The remaining models display further advancements, with the proposed HgbNet attaining the lowest RMSE and MAE, as well as the highest R2 scores in both datasets. The improvements   over the best metric from  other baselines in the two datasets are 5.2$\%$, 16.2$\%$, 0.6$\%$ and 19.5$\%$, 6.4$\%$, 1.9$\%$. Concerning the Data-GRU, ATTAIN, MIAM, and HiTANet methods, the latter two yield marginally better results than the former two, possibly due to their incorporation of various attention types.

 Table \ref{The evaluation matrix for anemia degree estimation.} compares model performance in predicting anemia degrees.   Generally, the results align with the hemoglobin  prediction  findings. LSTM and T-LSTM display limited capabilities, while other baselines exhibit improvement, with HgbNet showcasing the best performance. It achieves the highest weighted precision and weighted F1 scores in both datasets. Specifically, the improvements over the best results from other baselines amount to 2.8$\%$, 1.3$\%$, 0.2$\%$, and 1.2$\%$.
In conclusion, all baseline models improve in handling irregular time series compared to traditional LSTM methods. Our proposed HgbNet consistently outperforms existing methods for both tasks across the two datasets, showcasing its superior performance.

%我觉得要说一下时间的分布，然后给出time interval的分布直方图
\subsubsection{Irregular time interval analysis}
In practice, the interval between moments $T$ and $T+1$ is indeterminate (i.e., the patient's last diagnosis result could have been hours or days ago), and excessively large intervals may render predictions unreliable. We now aim to explore this issue further.

\begin{figure}[hbtp]
\centering
\subfigure[MIMIC III dataset]{
\includegraphics[width=9cm]{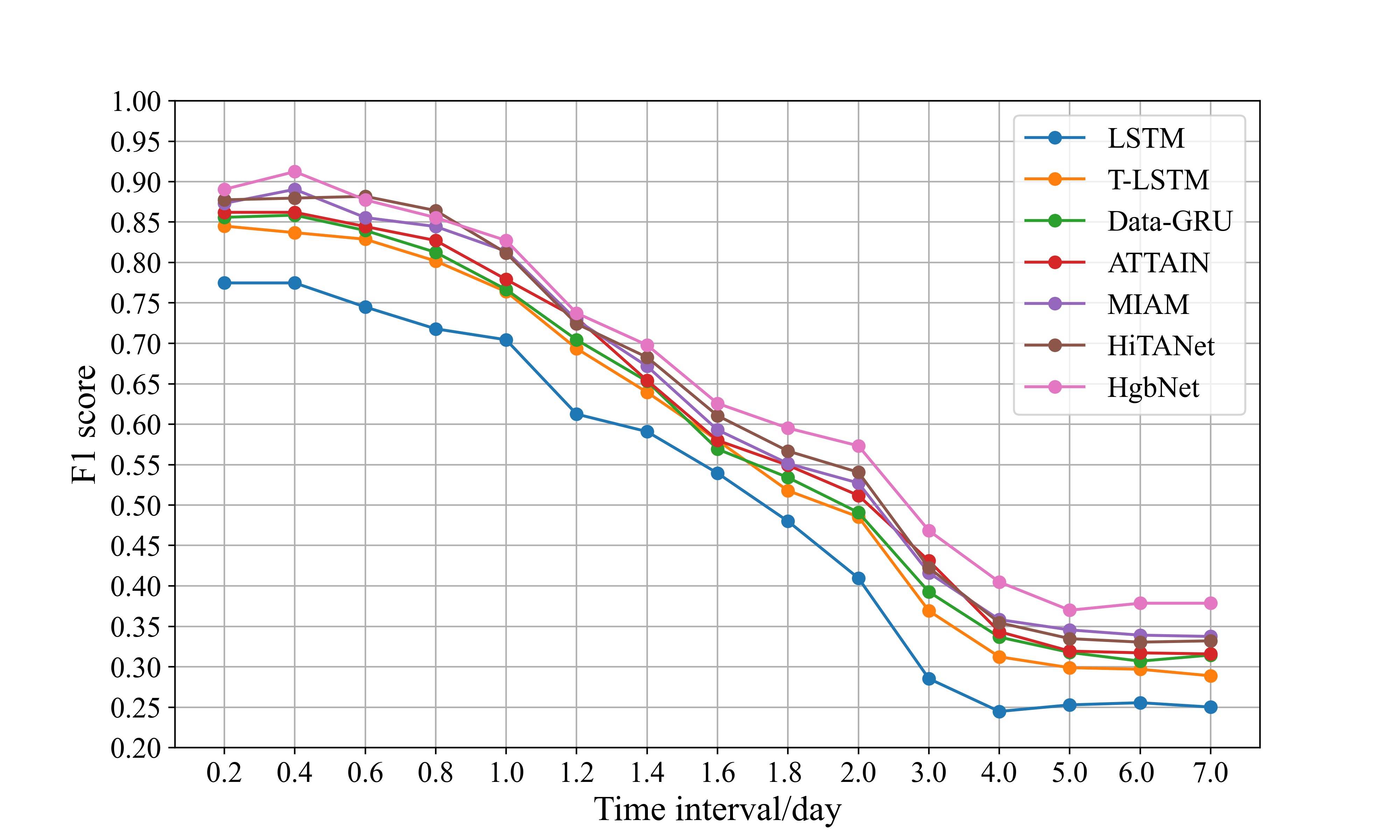}
%\caption{fig1}
}\hspace{-7mm}
\subfigure[eICU dataset]{
\includegraphics[width=9cm]{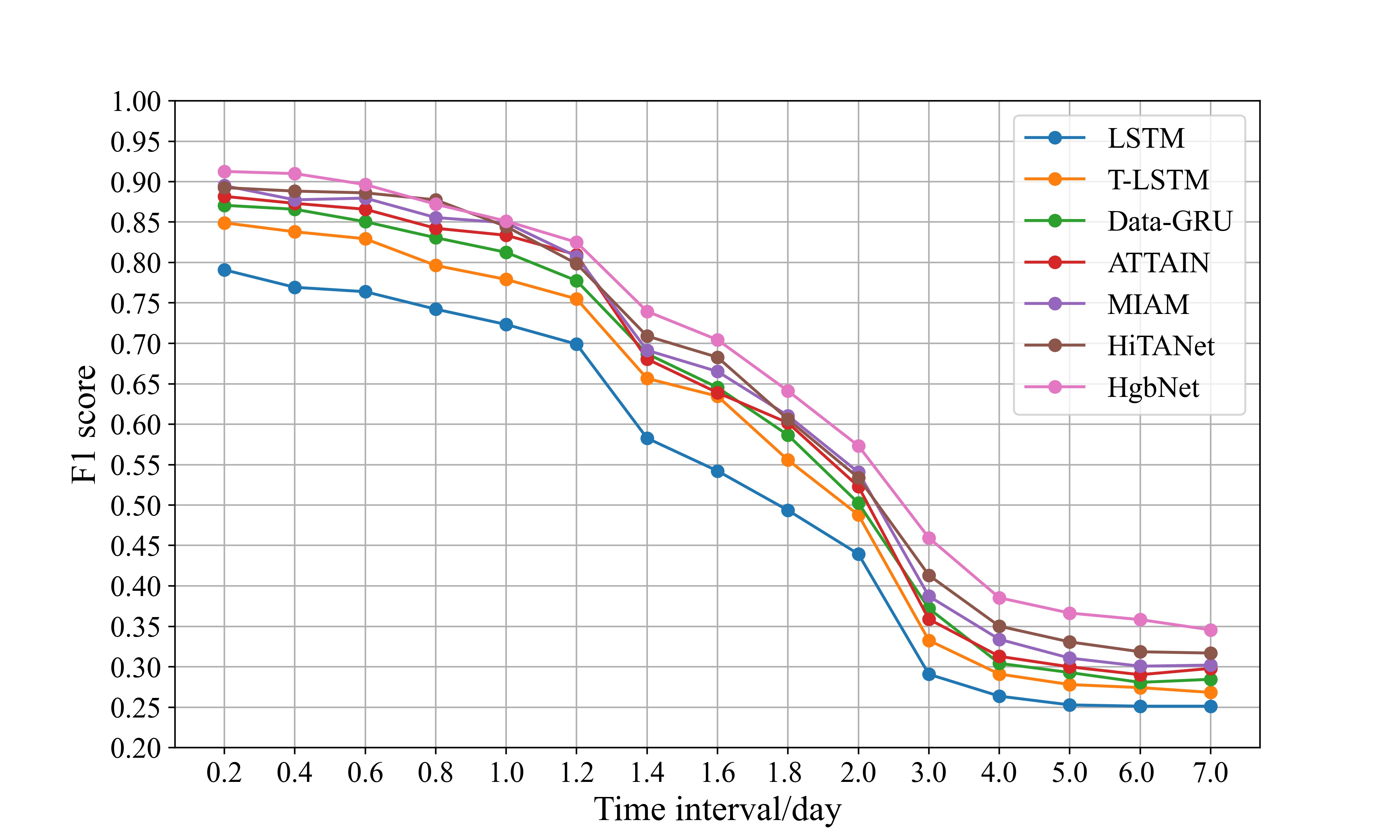}
%\caption{fig1}
}\hspace{-7mm}
\centering
\caption{The  anemia degree prediction results under irregular time intervals.}
\label{case1 anemia}
\end{figure}

Due to space constraints, we employ R2 scores and F1 scores as evaluation metrics for two tasks, respectively. The prediction results under irregular time intervals are presented in Fig. \ref{case1 hemoglobin} and Fig. \ref{case1 anemia}. The insights derived from Fig. \ref{case1 hemoglobin} are as follows:
%hemoglobin level的图
\begin{figure}[hbtp]
\centering
\subfigure[MIMIC III dataset]{
\includegraphics[width=9cm]{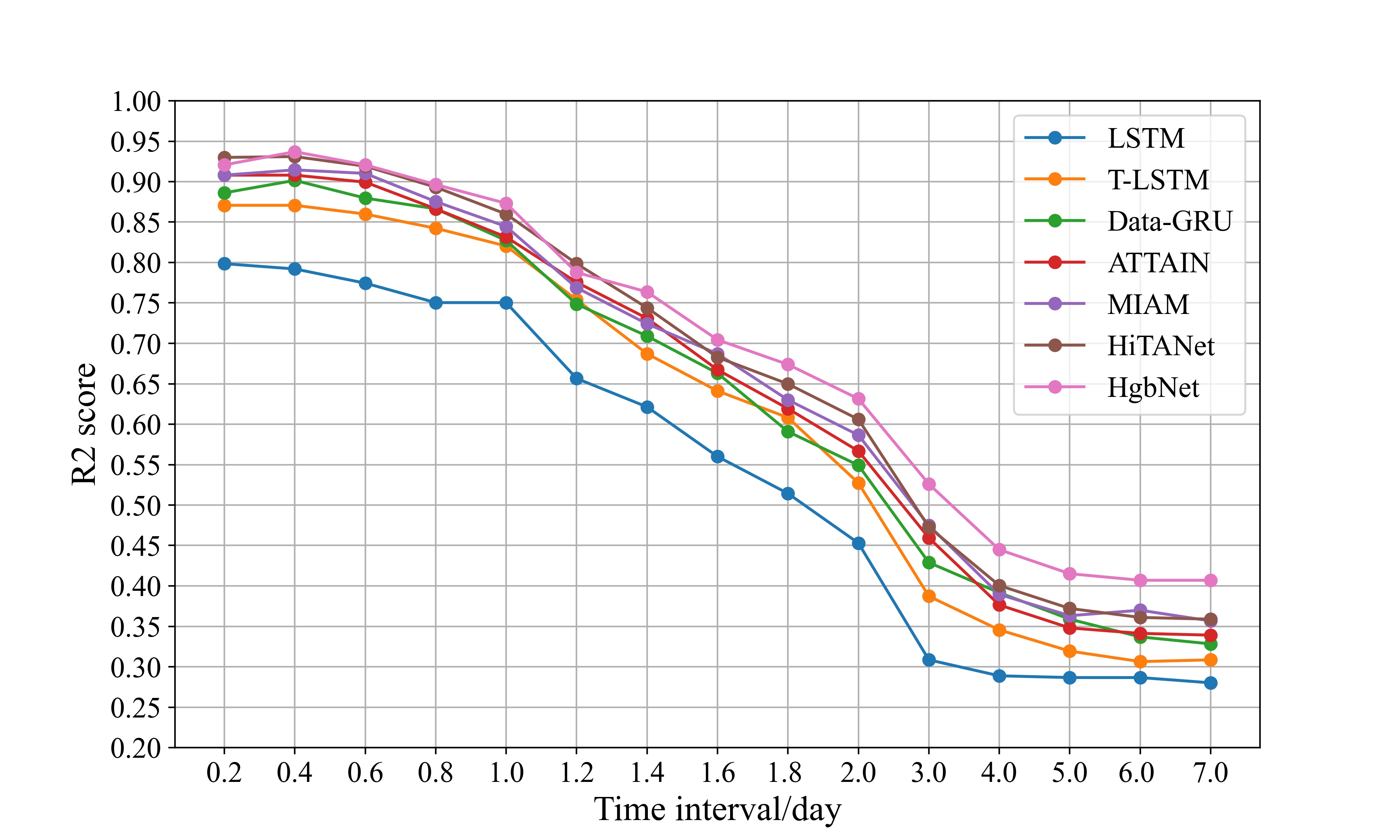}
%\caption{fig1}
}\hspace{-7mm}
\subfigure[eICU dataset]{
\includegraphics[width=9cm]{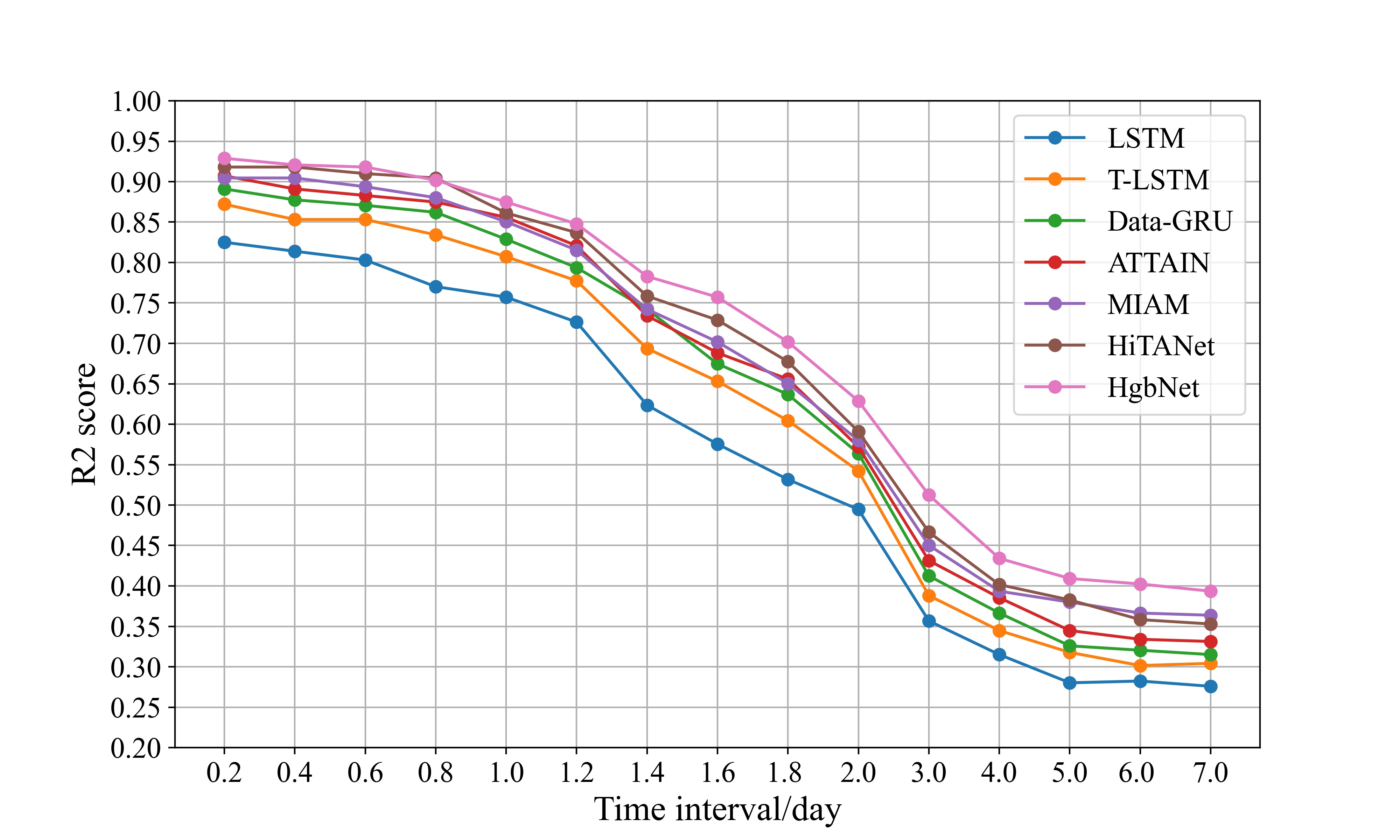}
%\caption{fig1}
}\hspace{-7mm}
\centering
\caption{The  hemoglobin level prediction results under irregular time intervals.}
\label{case1 hemoglobin}
\end{figure}

%anemia degree的图

    $\bullet$ For hemoglobin level prediction, all models exhibit a decline in performance as the forecasting time interval increases, with a non-uniform rate of decrease.   Specifically, in  the MIMIC III dataset, high R2 scores are  preserved within the interval of 0-1 day. Nearly all models achieve their peak R2 scores (0.937, 0.931, 0.915, 0.908, 0.902, 0.871, and 0.799 for each model) at the interval of 0.2 or 0.4 days. The R2 score decay rates for these models, from the 0.2-day to 1-day intervals, are 6.0$\%$, 5.8$\%$, 6.7$\%$, 8.4$\%$, 7.0$\%$, 7.5$\%$, and 5.2$\%$, respectively.  A notable drop in performance occurs between the interval of 1 and 3 days, with respective  drop rates of 58.9$\%$, 52.8$\%$, 48.2$\%$, 44.7$\%$, 43.8$\%$, 45.0$\%$ and 39.8$\%$, respectively. In time intervals of 3 to 7 days, the models exhibit low performance and gradually stabilize at a constant value. R2 scores at this point are between 0.25 and 0.4, suggesting poor model performance and unsuitability for practical applications. For the eICU dataset, a similar trend is observed,  with rapid model performance decay from 1.2 to 4 days, and stabilization thereafter.  The above phenomena can be attributed to 1) the majority  of the time intervals in both datasets being concentrated within 0-1.2 days, and 2) from a medical standpoint, patients with longer intervals present a high degree of uncertainty, rendering many physiological parameters unsuitable for prediction.
    
    $\bullet$ Our proposed ND-LSTM model consistently demonstrates nearly the highest performance across all time intervals, with evident advantages for intervals exceeding 1 day. This suggests that considering both local and global irregularity  is beneficial for longer interval predictions.  Conversely, the LSTM model exhibits the poorest performance across all time intervals, with the fastest decay to the minimum value.

The results of the anemia degree prediction experiment, depicted in
Fig. \ref{case1 anemia}, are  consistent with the findings of the hemoglobin level prediction experiment. All models attain peak performance at 0.2 or 0.4 days, with F1 scores ranging between 0.75 and 0.92. They exhibit a gradual performance decline from 0 to 1/1.2 days, followed by a sharp decrease from 1/1.2 to 3 days in both datasets. The proposed HgbNet demonstrates a marginal advantage from 0 to 1 day, which is further extended at intervals beyond 1 day. It is worth noting that the LSTM model yields the lowest F1 score, approximately 0.25, equivalent to a random guess.

In conclusion, our proposed HgbNet  displays outstanding performance across various time intervals.  However,  it is crucial to emphasize results within time intervals of less than 1 day, as performance deteriorates rapidly beyond this range.
\subsection{Use case 2}\label{HAHA}
In this case, we predict the patient's  hemoglobin level/anemia degree at moment $T+1$ by utilizing  data from time point 1 to time point $T$,  along  with selected  test values at $T+1$.   We choose the following non-invasive and easily accessible test results at moment $T+1$ (available in both datasets): Spo2, temperature, heart rate, respiratory rate and blood pressure. 

\begin{figure}[H]
\centering
\subfigure[MIMIC III dataset]{
\includegraphics[width=9cm]{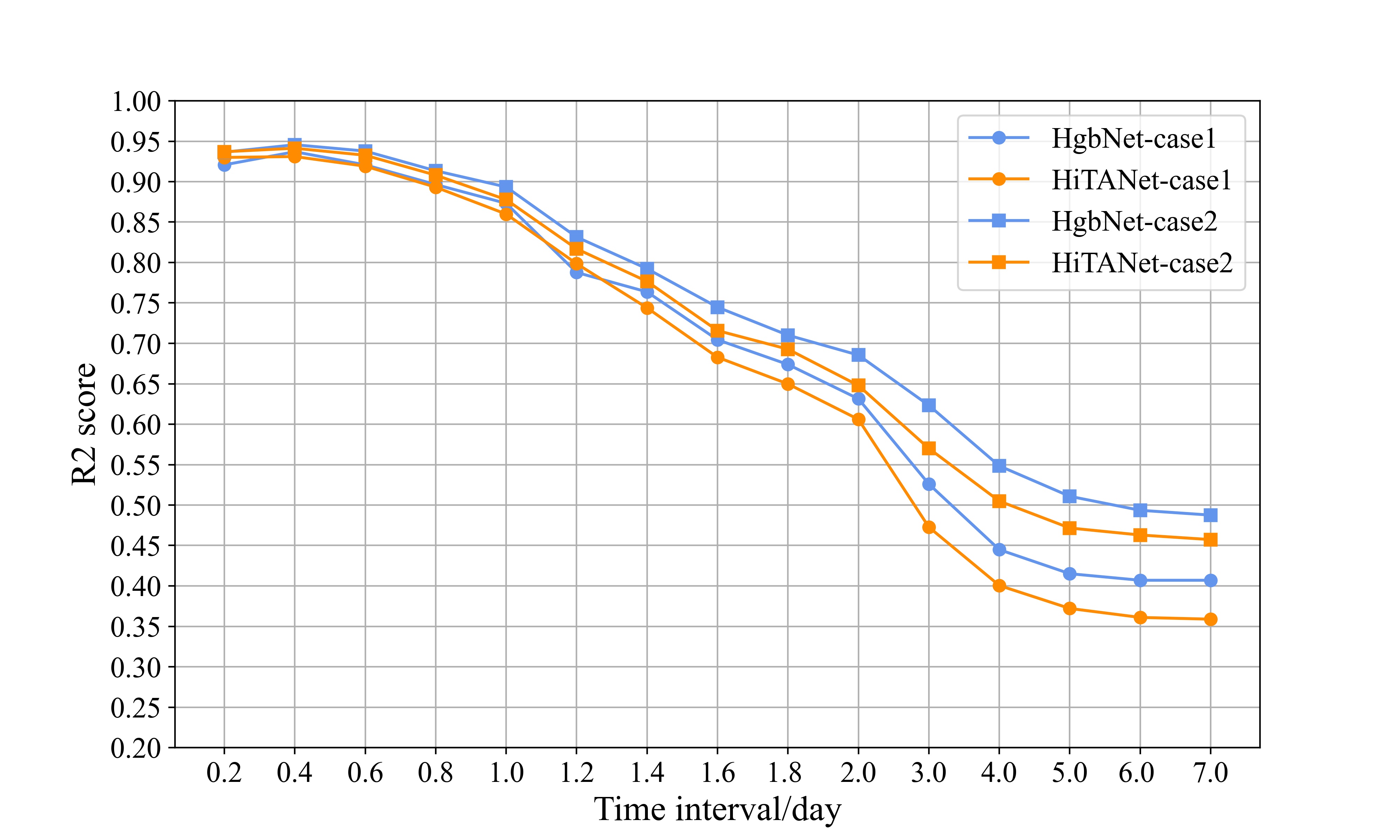}
%\caption{fig1}
}\hspace{-7mm}
\subfigure[eICU dataset]{
\includegraphics[width=9cm]{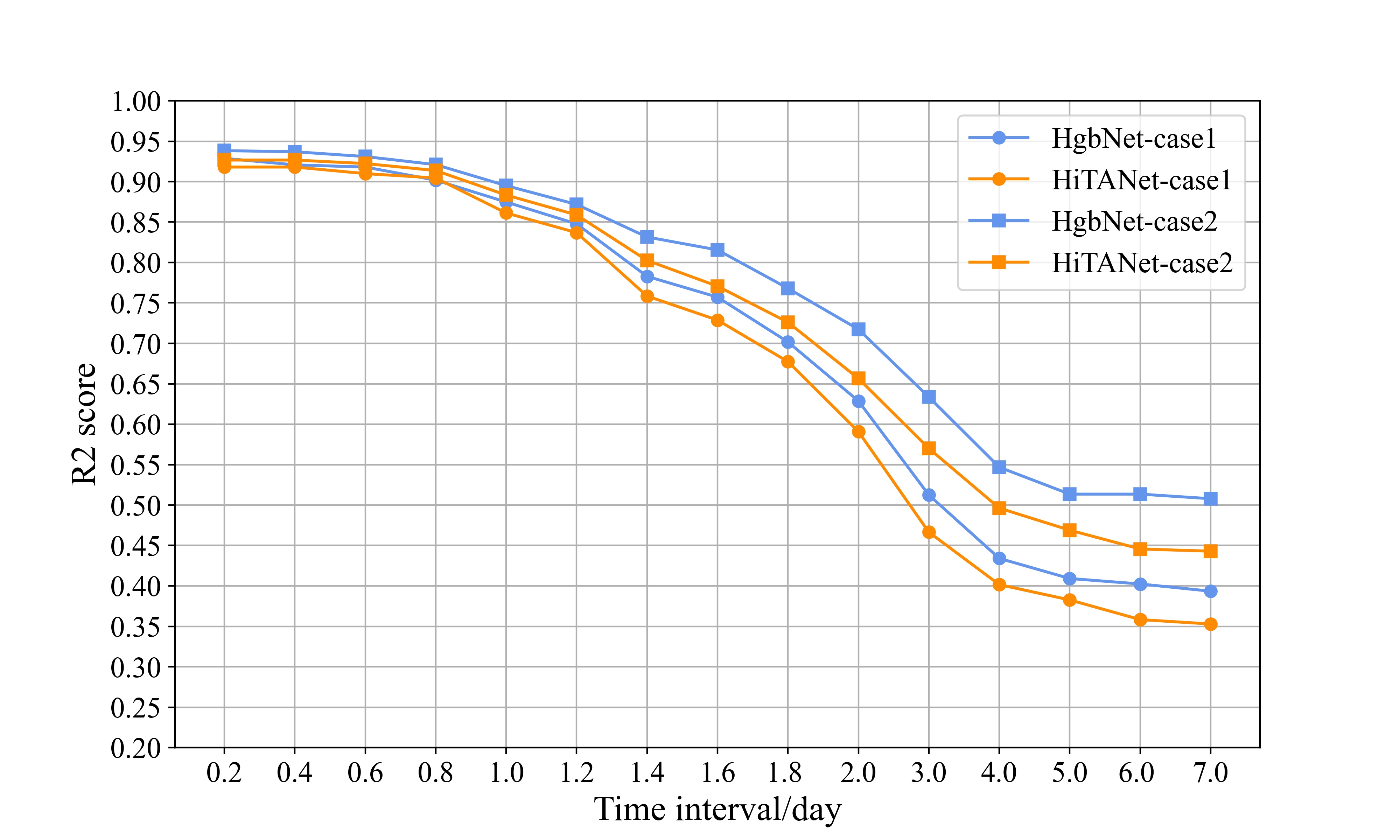}
%\caption{fig1}
}\hspace{-7mm}

\centering
\caption{The hemoglobin level prediction result at different time intervals of two datasets.}
\label{case2_hemoglobin}
\end{figure}

\begin{figure}[H]
\centering
\subfigure[MIMIC III dataset]{
\includegraphics[width=9cm]{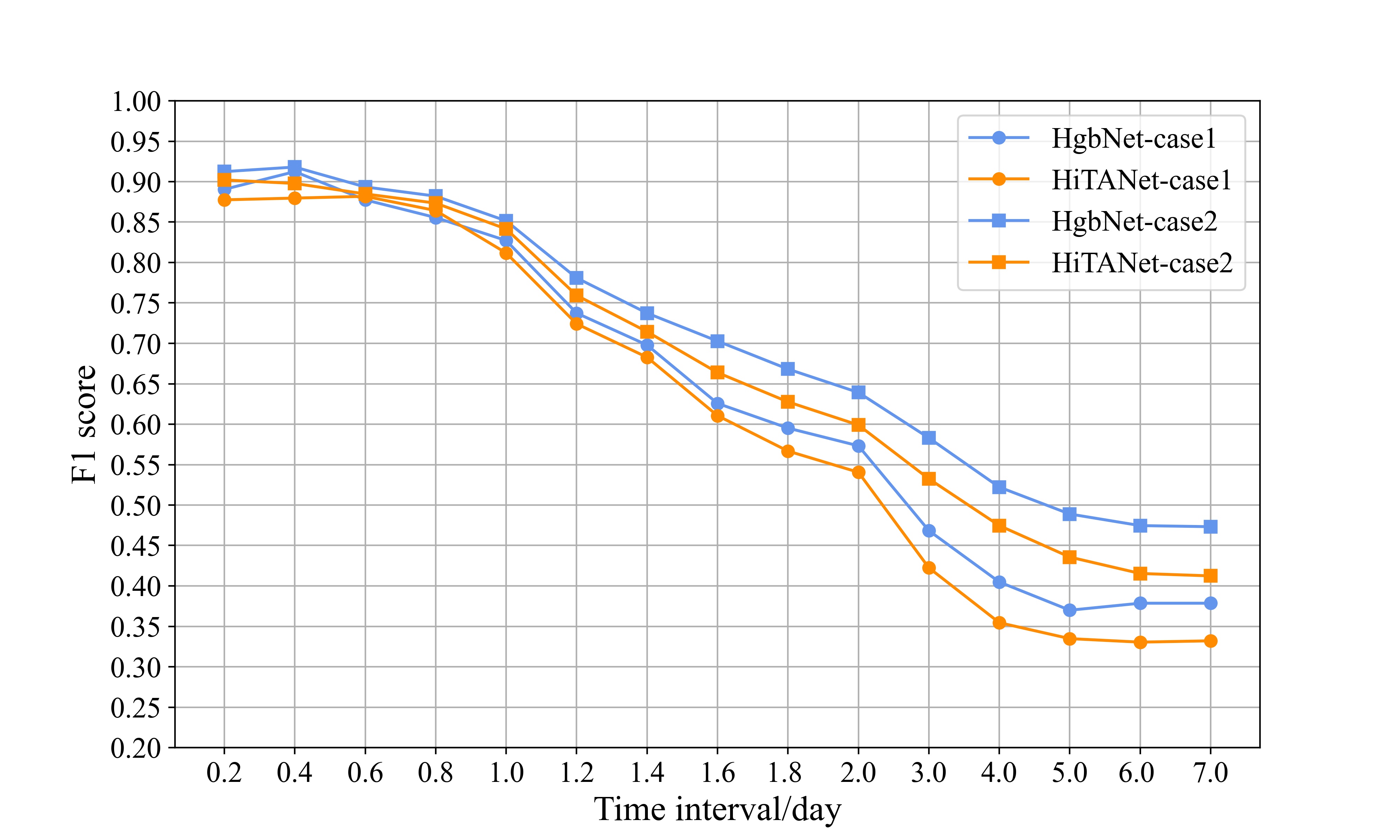}
%\caption{fig1}
}\hspace{-7mm}
\subfigure[eICU dataset]{
\includegraphics[width=9cm]{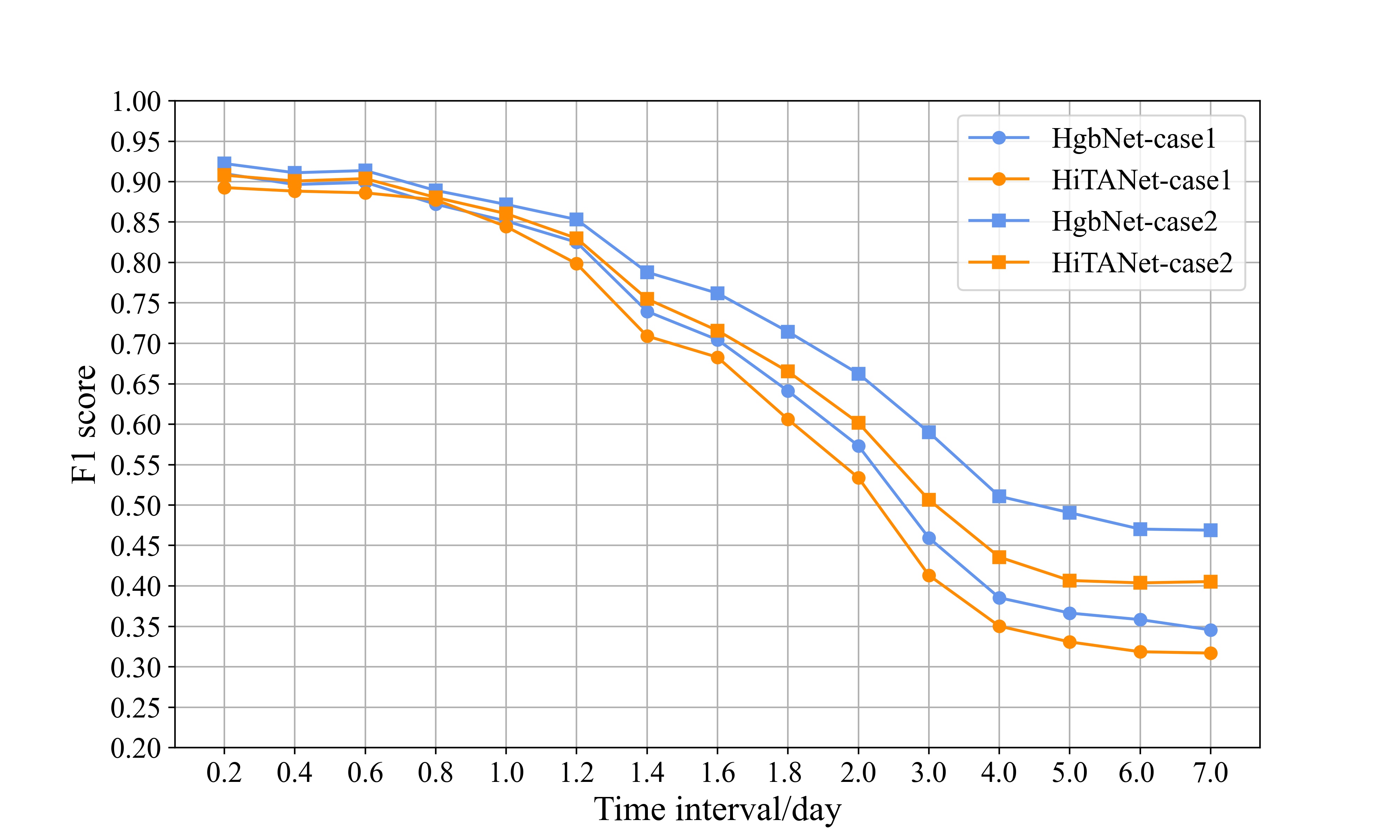}
%\caption{fig1}
}\hspace{-7mm}
\centering
\caption{The anemia degree prediction result at different time intervals of two datasets.}
\label{case2_anemia}
\end{figure}

The overall results of the hemoglobin level  prediction  and the anemia degree prediction experiment  are presented  in Table \ref{Overall_hemoglobin} and  \ref{Overall_anemia}, respectively. For experiments analyzing irregular time intervals, we selected the two top-performing models, HgbNet and HiTANet, for comparison. Their experimental outcomes in use case 1 versus use case 2 are illustrated in Fig. \ref{case2_hemoglobin} and Fig. \ref{case2_anemia}.

%hemoglobin level case2的表
\begin{table*}[htbp]\footnotesize
	\centering
	\setlength{\tabcolsep}{2mm}
	\caption{The evaluation matrix for hemoglobin level estimation.}\label{Overall_hemoglobin}
	\begin{tabular}{ccccccc}
		\hline
		Dataset  & \multicolumn{3}{c}{MIMIC III} & \multicolumn{3}{c}{eICU}\\ \hline
		\diagbox[dir=NW,height=2em,trim=r]{\footnotesize Model}{\footnotesize Metrics}             & RMSE     & MAE     & R2 score     & RMSE     & MAE     & R2 score     \\ \hline
LSTM       & 2.042 $\pm$\ 0.014  &1.648  $\pm$\ 0.012& 0.751 $\pm$\  0.010  & 1.372 $\pm$\ 0.013 & 1.155 $\pm$\ 0.012 &0.774 $\pm$\ 0.009  \\
T-LSTM     & 1.541 $\pm$\ 0.011 &1.143  $\pm$\ 0.010& 0.822 $\pm$\  0.006  & 0.997 $\pm$\ 0.009& 0.769 $\pm$\  0.010&0.839 $\pm$\ 0.006  \\
Data-GRU   & 1.112 $\pm$\ 0.009  &0.873  $\pm$\ 0.006& 0.869 $\pm$\ 0.004   & 0.853 $\pm$\ 0.007& 0.682 $\pm$\  0.009&0.867 $\pm$\ 0.004  \\
ATTAIN     & 1.029 $\pm$\ 0.008  &0.824  $\pm$\ 0.008& 0.862 $\pm$\  0.005  & 0.894 $\pm$\ 0.009& 0.652 $\pm$\ 0.007 &0.850 $\pm$\  0.002 \\
MIAM       & 0.838 $\pm$\  0.008&0.665  $\pm$\ 0.007& 0.884 $\pm$\  0.003  & 0.729 $\pm$\ 0.008& 0.549 $\pm$\  0.007 &0.879 $\pm$\ 0.003  \\ 
HiTANet    & 0.794 $\pm$\ 0.007 &0.610  $\pm$\ 0.008& 0.893 $\pm$\  0.004  & 0.774 $\pm$\ 0.007& 0.591 $\pm$\ 0.008 &0.873 $\pm$\ 0.004  \\ 
HgbNet  & \textbf{0.716} $\pm$\ 0.006  &\textbf{0.561} $\pm$\ 0.007  & \textbf{0.917} $\pm$\  0.003    &  \textbf{0.612} $\pm$\ 0.007  & \textbf{0.597} $\pm$\ 0.007  &\textbf{0.901} $\pm$\ 0.002 \\ 
%elcu的r2score会低一些但是rmse mae会好一些，因为杂点少,这里考虑要不要把这两个数据集的rmse和mae给换一下或者是把rmse和mae再调大一些,我决定把rmse和mae都调大1.3倍
\hline
	\end{tabular}
\end{table*}

%anemia degree case2 的表
\begin{table*}[htbp]\footnotesize
	\centering
	\setlength{\tabcolsep}{2mm}
	\caption{The evaluation matrix for anemia degree estimation.}\label{Overall_anemia}
	\begin{tabular}{ccccccc}
		\hline
		Dataset & \multicolumn{3}{c}{MIMIC III} & \multicolumn{3}{c}{eICU}\\ \hline
		\diagbox[dir=NW,height=2em,trim=r]{\footnotesize Model}{\footnotesize Metrics}

  & \tabincell{c}{Weighted\\Precision}  & \tabincell{c}{Weighted\\Recall}  &\tabincell{c}{Weighted\\F1 score}  & \tabincell{c}{Weighted\\Precision}  & \tabincell{c}{Weighted\\Recall}& \tabincell{c}{Weighted\\F1 score}   \\ \hline
  
LSTM     &0.694 $\pm$\ 0.011& 0. 734  $\pm$\  0.012 &  0.675 $\pm$\ 0.011  &0.749 $\pm$\ 0.013      &0.744 $\pm$\ 0.012 &0.729 $\pm$ 0.012\ \\ 
T-LSTM   & 0.827 $\pm$\ 0.008& 0.825  $\pm$\ 0.008 &0.823 $\pm$\  0.007  &0.824   $\pm$\ 0.008      &0.826 $\pm$\ 0.008 &0.826 $\pm$ 0.009\ \\      
Data-GRU & 0.849 $\pm$\ 0.006& 0.857 $\pm$\ 0.007 &  0.853 $\pm$\  0.006  &0.848  $\pm$\ 0.005      &0.844 $\pm$\ 0.006 &0.842  $\pm$ 0.006\\\ 
ATTAIN   &0.863 $\pm$\ 0.005 &0.857 $\pm$\ 0.005 &0.857 $\pm$\  0.005     &0.855  $\pm$\ 0.005      &0.848 $\pm$\ 0.006 &0.849 $\pm$ 0.004\ \\
MIAM       & 0.874 $\pm$\ 0.007& 0.875 $\pm$\ 0.008 &0.872 $\pm$\  0.005  &0.862  $\pm$\ 0.006      &0.863 $\pm$\ 0.005&0.858 $\pm$ 0.005\ \\
HiTANet    & 0.887 $\pm$\ 0.005& 0.891  $\pm$\ 0.006&0.879 $\pm$\  0.006  &0.864  $\pm$\  0.004     &0.867 $\pm$\ 0.004&0.864 $\pm$ 0.005\ \\
HgbNet  &\textbf{0.900} $\pm$\ 0.006  &\textbf{0.895}   $\pm$\ 0.006 &\textbf{0.897 }  $\pm$\  0.005  &\textbf{0.890} $\pm$\  0.005  &\textbf{0.890} $\pm$\ 0.004 &\textbf{0.889} $\pm$ 0.004\ \\  
\hline
	\end{tabular}
\end{table*}

%hemoglobin level case2的图

%anemia degree case2的图

Table \ref{Overall_hemoglobin} reveals that the prediction results for hemoglobin levels are improved across all models compared to use case 1. For example, the R2 score has increased by 9.6$\%$, 2.1$\%$, 3.1$\%$, 3.1$\%$, 4.1$\%$, 3.6$\%$, and 5.8$\%$ for the MIMIC III dataset, and 7.4$\%$, 3.6$\%$, 5.5$\%$, 2.7$\%$, 4.0$\%$, 3.6$\%$, and 4.6$\%$ for the eICU dataset.  As depicted  in Fig. \ref{case2_hemoglobin}, similar trends are observed in both datasets: incorporating partial test results at  $T+1$ doesn't notably enhance results with time intervals of less than 1 day. However, more substantial improvements occur when time intervals exceed 1 day. For instance, the average R2 score improvements over every five consecutive time intervals for HgbNet are 1.6$\%$, 4.3$\%$, 12.8$\%$ for the MIMIC III dataset and 1.6$\%$, 4.6$\%$, 17.1$\%$ for the eICU dataset. We attribute this phenomenon to two factors: 1) as previously noted, most time interval samples are concentrated within  0-1 day, making it challenging to further optimize and improve results; 2) the tests selected for time point $T+1$ are non-invasive and rapid, with limited contribution to predicting hemoglobin concentration.

The anemia degree prediction experiment results,  displayed in Table  \ref{Overall_anemia} and Fig. \ref{case2_anemia}, exhibit a similar trend.  For instance, considering the F1 score, the performance of all models improves by  5.63$\%$, 3.0$\%$, 2.9$\%$, 4.9$\%$, 4.3$\%$, 4.1$\%$, 5.2$\%$ for the MIMIC III dataset and 8.6$\%$, 7.3$\%$, 2.2$\%$, 4.2$\%$, 2.5$\%$, 4.0$\%$, 4.7$\%$ for the eICU dataset. In the analysis of irregular time intervals, a noticeable enhancement in the performance is observed when the time interval is greater than 1 day. The average F1 score improvements over every five consecutive time intervals are 2.0$\%$, 6.4$\%$, 18.2$\%$ for the MIMIC III dataset and 1.6$\%$, 5.0$\%$, 20.3$\%$ for the eICU dataset. 

In summary,  the performance of the HgbNet  on both tasks and use cases has been  improved by providing certain test values at the moment $T+1$.

\subsection{Ablation study}
\textcolor{black}{To investigate the performance of the main components (i.e. NanDense layer, global attention, feature-level attention and label-level attention) in the proposed HgbNet, we designed ablation experiments. In these ablation experiments, we implement the hemoglobin level and  anemia degree prediction under use case 1. R2 score and F1 score are involved as the metric. At the same time, we compare the performance of NanDense with other popular imputation methods (i.e. mean value imputation (IM-mean) \cite{zhang2016missing} and GAN imputation (IM-GAN) \cite{luo2018multivariate}).  The results of the ablation study are shown in Table \ref{Ablation study}.}

\begin{table*}[h]
 	\centering
        \small
 	\setlength{\tabcolsep}{2mm}
 	\caption{\textcolor{black}{Performance of the different module combinations in HgbNet.  ND, AT1, AT2 and AT3 refer to the NanDense layer, general attention, feature-level attention and label-level attention, respectively. Other modules not mentioned (as described in Sec \ref{HgbNet}) are applied in all combinations by default.}}\label{Ablation study}
 	\begin{tabular}{ccccc}
  		\hline
		Dataset  & \multicolumn{2}{c}{MIMIC III} & \multicolumn{2}{c}{eICU}\\ \hline
		\diagbox[dir=NW,height=2em,trim=r]{\footnotesize Module}{\footnotesize Metrics}      
 		&\tabincell{c}{R2\\ score } 
 		&\tabincell{c}{F1 \\score }
    	&\tabincell{c}{R2\\ score } 
 		&\tabincell{c}{F1\\ score }
   \\ \hline
 		
 \tabincell{c}{ND}     & 0.821 $\pm$\ 0.008 &0.813 $\pm$\ 0.011 & 0.816 $\pm$\ 0.009 &0.794 $\pm$\ 0.012\\
  \tabincell{c}{ND+AT1}& 0.848 $\pm$\ 0.005 &0.842 $\pm$\ 0.007 & 0.837 $\pm$\ 0.005 &0.820 $\pm$\ 0.007\\  %最差
 \tabincell{c}{ND+AT2} &0.859 $\pm$\ 0.004 &0.847 $\pm$\ 0.005 & 0.852 $\pm$\ 0.003 &0.836 $\pm$\ 0.005\\ %最好
 \tabincell{c}{ND+AT3} & 0.851 $\pm$\ 0.004&0.842 $\pm$\ 0.006 & 0.840 $\pm$\ 0.003 &0.829 $\pm$\ 0.005\\ % 其次
  \tabincell{c}{IM-mean+AT1+AT2+AT3} & 0.855 $\pm$\ 0.005 &0.837 $\pm$\ 0.007 & 0.844 $\pm$\ 0.003 &0.828 $\pm$\ 0.005\\
   \tabincell{c}{IM-GAN+AT1+AT2+AT3} & 0.858 $\pm$\ 0.003 &0.840 $\pm$\ 0.006 & 0.839 $\pm$\ 0.004 &0.833 $\pm$\ 0.006\\
 \tabincell{c}{ND+AT1+AT2+AT3} & \textbf{0.867} $\pm$\ 0.003  &\textbf{0.855} $\pm$\ 0.005 &\textbf{0.861} $\pm$\ 0.003  &\textbf{0.843} $\pm$\ 0.005 \\
 \hline
 	\end{tabular}
 \end{table*}

\textcolor{black}{It can be seen from Table \ref{Ablation study} that  the absence of an attention module leads to the poorest performance, marginally surpassing that of T-LSTM. Among the three examined attention modules, feature-level attention markedly enhances performance. This improvement is primarily attributed to the mechanism's detailed consideration of each test item's unique irregularities. Conversely, general attention yields a relatively modest enhancement, stemming from its failure to account for the temporal decay in these irregularities. Furthermore, we observe a notable advancement in performance with the NanDense layer approach for handling missing values, compared to the mean/GAN imputation method. This superiority likely arises from the NanDense layer's strategy of not making assumptions in predicting missing values, thereby reducing the risk of introducing outliers.}

%% file: SPLITEE/chapters/Conclusion.tex
\section{Conclusion}
Predicting hemoglobin levels and anemia severity from EHR data is a compelling yet highly challenging task.  In this study, we introduce  HgbNet, which addresses the limitations of existing methods. Specifically, we employ a NanDense layer and a missing indicator to manage missing values in EHR data. We also implement feature-level and label-level attention, along with local attention, to tackle the issue of irregular time intervals.   In use case 1, we achieve R2 scores of 0.867 and 0.861 for hemoglobin level prediction, and F1 scores of 0.855 and 0.843 for anemia severity prediction across both datasets, which surpass  the performance of  the current SOTA methods.

We also evaluate the impact of irregular time intervals on prediction results: we find that HgbNet performs well for forecasting periods shorter than 1 day, but its performance declines rapidly for longer intervals. In use case 2, HgbNet's performance is enhanced by introducing test values of Spo2,
temperature, heart rate, respiratory rate and  blood pressure at moment $T+1$, resulting in improvements of 5.8$\%$ and 3.6$\%$ for R2 scores, and 5.2$\%$ and 4.7$\%$ for F1 scores.  
% Furthermore, the rate of serious misdiagnosis is significantly reduced. 
These findings suggest that HgbNet holds substantial potential for EHR-based hemoglobin level and anemia severity estimation.
 
 From a clinical perspective, incorporating HgbNet into existing biochemistry systems or EHR platforms could serve as a valuable safety net, preventing overlooked cases of anemia or alerting clinicians to trends in hemoglobin levels. For conditions like thalassemia, where frequent monitoring and blood transfusions are essential for patients, HgbNet could be integrated into care pathways. The inclusion of non-invasive sensor data could further enhance patient compliance with monitoring. The primary objective of these systems would be to directly improve health outcomes by reducing the need for testing and minimizing physician error when reviewing multiple patients' hemoglobin values in a single session.

\textcolor{black}{We have showcased the effectiveness of incorporating non-invasive features in enhancing prediction outcomes. In our future research, we plan to delve deeper into specific methodologies, such as attention analysis, to determine the most suitable non-invasive tests for selection or even to design specialized sensors.}